\documentclass[11pt]{article}

\usepackage[final]{acl}

\usepackage{cuted}
\usepackage{caption}

\usepackage{times}
\usepackage{latexsym}

\usepackage[T1]{fontenc}

\usepackage[utf8]{inputenc}

\usepackage{microtype}

\usepackage{inconsolata}

\usepackage{graphicx}
\usepackage{amsmath}
\usepackage{amssymb}
\usepackage{booktabs}
\usepackage{threeparttable}
\usepackage{tikz}
\usepackage{wrapfig}
\usetikzlibrary{positioning, arrows.meta, shapes.geometric, fit, backgrounds, calc}
\usepackage{xcolor}
\definecolor{matcha}{RGB}{107,200,170}

\definecolor{opRename}{HTML}{2C5F8A}
\definecolor{opMerge} {HTML}{A8601B}
\definecolor{opSplit} {HTML}{6B4E9B}
\definecolor{opKeep}  {HTML}{808080}
\definecolor{opgray}  {HTML}{595959}
\newcommand{\optag}[2]{{\setlength{\fboxsep}{1.7pt}%
  \colorbox{#1!10}{\scriptsize\textsc{\textcolor{#1}{#2}}}}\ }
\newcommand{\oprename}{\optag{opRename}{rename}}
\newcommand{\opmerge} {\optag{opMerge}{merge}}
\newcommand{\opsplit} {\optag{opSplit}{split}}
\newcommand{\opkeep}  {{\setlength{\fboxsep}{1.7pt}%
  \colorbox{opKeep!10}{\scriptsize\textsc{\textcolor{opKeep}{unchanged}}}}}
\newcommand{\sub}[1]{\textcolor{opgray}{+}~#1}
\usepackage{fontawesome5}
\usepackage{tabularx}
\usepackage{fancyvrb}
\usepackage{fvextra}
\usepackage{colortbl}
\usepackage{tcolorbox}
\tcbuselibrary{skins, breakable, listings}
\usepackage{listings}
\usepackage{stfloats}
\usepackage{placeins}
\usepackage{cuted}
\usepackage{float}
\usepackage{enumitem}
\usepackage{multirow}
\usepackage{afterpage}
\usepackage{flafter}
\usepackage{xspace}
\usepackage{fontawesome5}

\newcommand{\peeragent}{\emph{Peer-Debriefing Agent}\xspace}
\newcommand{\codingagent}{\emph{Hierarchical Coding Agent}\xspace}
\newcommand{\systemname}{\texttt{Agent-as-Peer-Debriefer}\xspace}

\newif\ifshowcomments
\showcommentsfalse  

\ifshowcomments
  \newcommand{\gao}[1]{\textcolor{blue}{[\textbf{Gao:} #1]}}
  \newcommand{\fan}[1]{\textcolor{red}{[\textbf{Fan:} #1]}}
  \newcommand{\zhimin}[1]{\textcolor{cyan}{[\textbf{zhimin:} #1]}}
  \newcommand{\kun}[1]{\textcolor{purple}{[\textbf{kun:} #1]}}
\else
  \newcommand{\gao}[1]{}
  \newcommand{\fan}[1]{}
  \newcommand{\zhimin}[1]{}
  \newcommand{\kun}[1]{}
\fi

\title{Agent-as-Peer-Debriefer: A Multi-Agent Framework with Perspective-Based Refinement for Qualitative Analysis}

\author{
  Zhimin Lin$^{1}$,
  Kun Cheng$^{1}$,
  Zhiyao Shu$^{2}$,
  Junhua Fang$^{1}$,
  Juntao Li$^{1}$, 
  Fan Bai$^{3}$, 
  Jie Gao$^{3}$\thanks{\; Corresponding author.} \\
  $^{1}$Soochow University,
  $^{2}$George Mason University,
  $^{3}$Johns Hopkins University\\
  \texttt{linzhimin327@gmail.com} \hspace{0.5cm}
  \texttt{zshu2@gmu.edu} \hspace{0.5cm}
  \texttt{jgao77@jh.edu}\\
  \faGithub\ \url{https://github.com/NenRinCake/Agent-as-Peer-Debriefer}
}

\begin{document}
\maketitle
\begin{abstract}

Large language models (LLMs) are increasingly used for qualitative data analysis (QDA), yet their outputs often miss the depth and nuance of human analysis. We argue this gap reflects a missing credibility practice from human QDA: \textit{peer debriefing}, in which an analyst seeks feedback from a \textit{disinterested peer} and uses it to refine their coding. To bring this practice into LLM-assisted QDA, we propose \systemname, a multi-agent QDA framework that builds \textit{peer debriefing} into key coding steps. In our framework, a \codingagent follows the standard QDA process to generate codes, sub-themes, and themes, along with self-explanations and reflection memos. It then shares these outputs with three \peeragent{}s, each applying a distinct analytical perspective (Theory-Driven, Data-Driven, or Applied) and refining the codes by keeping, renaming, reassigning, merging, or splitting them. These perspectives are drawn from established human QDA practices that generalize across domains and datasets.
To evaluate the framework, we test it on three datasets across two domains with three LLMs, measuring semantic similarity to human-annotated codes. Across all settings, perspective-based, peer-debriefing refinement aligns more closely with human codes than a single-LLM baseline, and an ablation further shows the gain is not merely from additional refinement. The three perspectives also produce distinct trade-offs, showing that the choice of perspective is a meaningful and controllable design decision. More broadly, these findings suggest that simulating \textit{peer debriefing} with explicit perspectives is a promising route to more credible LLM-assisted QDA.
\end{abstract}

\section{Introduction}

\begin{figure}[t]
  \includegraphics[width=\columnwidth]{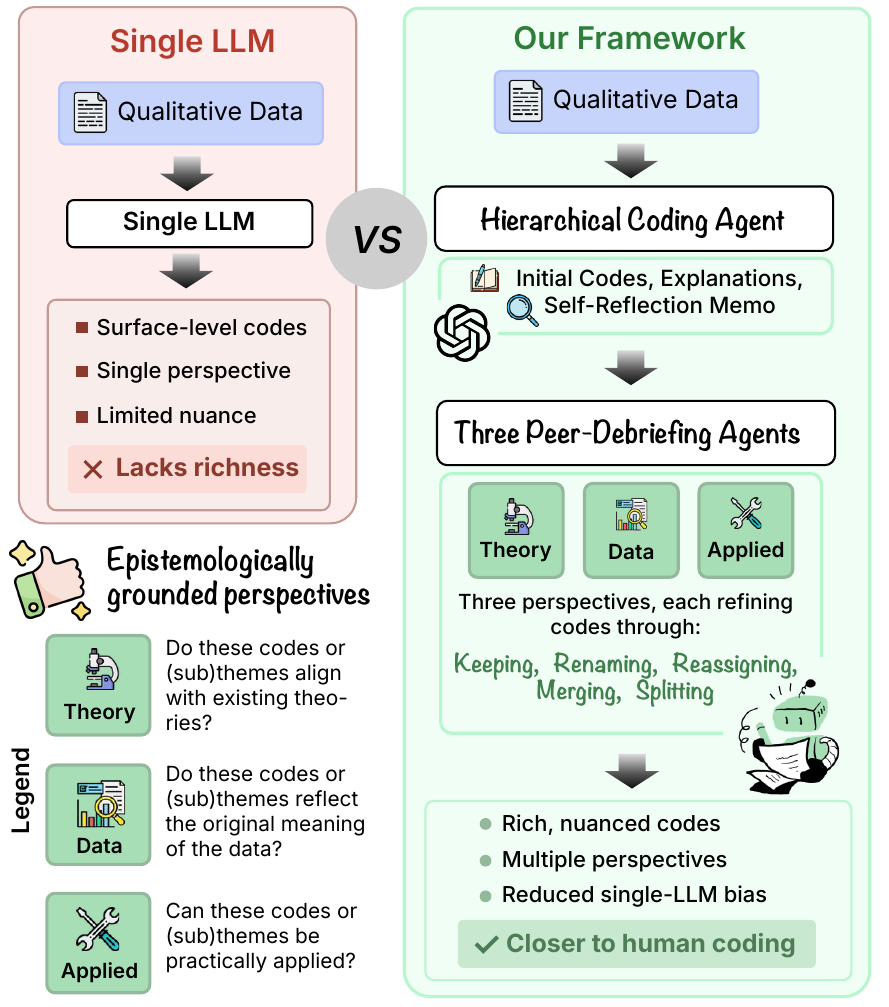}
  \caption{Motivation: A single LLM tends to produce surface-level codes. Our framework simulates peer debriefing: a \codingagent shares its initial codes, explanations, and self-reflection memo with \peeragent{}s to seek refinement feedback from them. The resulting codes are richer, more nuanced, and closer to human coding.}
  \label{fig:rewriting_on_difficult}
\end{figure}

Large language models (LLMs) are increasingly used to assist qualitative data analysis (QDA), from generating initial codes to synthesizing themes \cite{dai-etal-2023-llm, parfenova-etal-2024-automating, zhong-etal-2025-hicode}. Yet LLM-generated codes often lack the richness and nuance that human coding teams produce \cite{retkowski-etal-2025-ai, parfenova-etal-2025-text}. A key reason is that human teams do not rely on a single analyst's interpretation \cite{richards2018practical, gao2023coaicoder, gao2024using}. Instead, they use procedures like \textit{peer debriefing} to incorporate multiple perspectives and challenge initial readings of the data \cite{lincoln1985naturalistic, spall1998peer, barber2009conscience}.

\textit{Peer debriefing} is a standard practice in qualitative research: a researcher consults several colleagues with different backgrounds, often independently, to get feedback that helps refine their interpretation. These \textit{disinterested peers} are expected to understand the research context but have no personal stake in the outcome \cite{hail2011peer}. This process enriches the initial interpretation with nuances a single analyst would miss, and counters the blind spots that come from working alone.

In this work, we ask, \textbf{what if we bring peer debriefing into LLM-assisted QDA?} Can it bring codes from a single LLM closer to those produced by human teams? 
We propose \systemname, a multi-agent framework that simulates this: After generating initial coding (i.e., initial codes, subthemes, and themes), the \codingagent shares its output, explanation and self-reflection memos to three independent \peeragent{}s, who refine initial coding from different perspectives. Based on this, we investigate two research questions (RQs):
\begin{itemize}
    \item \textbf{RQ1:} Does \textit{peer debriefing} improve AI coding quality?
    \item \textbf{RQ2:} How do different analytical perspectives affect coding outcomes?
\end{itemize}

We compare codes refined under each perspective against a single-LLM baseline (no refinement), and an ablation study of self-refinement without perspective. For \textbf{RQ1}, we find that across all models and datasets, the best-performing perspective raises the match rate with human codes above the no-refinement baseline (e.g., GPT-5 on ScrumRQ1: 23.6\%$\to$46.9\%), and exceeds Self-Refinement on the same setting (32.3\%). For \textbf{RQ2}, we find that each perspective produces a distinct trade-off: Data-Driven maximizes recall of human codes (GPT-5 on ScrumRQ1: 70.6\%$\to$88.2\%), Theory-Driven and Applied maximize match rate (up to 46.9\%), and the no-refinement baseline keeps the most code diversity (25.9\% vs.\ 11--18\% under perspectives).
We summarize our contributions below:
\begin{itemize}
    \item We propose \systemname, a multi-agent framework that brings \textit{peer debriefing} into LLM-assisted QDA: a \codingagent shares its codes, explanations, and memos with three \peeragent{}s, each refining the initial coding from a distinct analytical perspective.
    \item We show that \textit{peer debriefing} consistently improves coding quality across datasets, domains, and LLMs, and that each perspective produces a distinct trade-off, making perspective choice a meaningful and controllable design decision in LLM-assisted QDA.
\end{itemize}

\section{Related Work}

\subsection{LLM-assisted QDA}

Computational tools have long supported qualitative coding and corpus reading~\cite{nelson2020computational, worden2017debates, evans2024algorithmic}. LLMs are the most recent of these tools, and NLP and HCI work has applied them to coding~\cite{than2025updating, schroeder2025large}.
Examples include interactive coding interfaces \cite{rietz2021cody, gebreegziabher2023patat}, deductive coding against a fixed codebook \cite{xiao2023supporting, tai2024examination}, inductive code and theme generation \cite{gao2024using, than2025updating}, and broader studies of LLMs in qualitative work \cite{gao2023coaicoder, bano2023large, hamilton2023exploring}. A recurring finding is that LLM-generated codes often differ from those of human coders. For example, Xiao et al.\ (\citeyear{xiao2023supporting}) found only fair to substantial agreement between GPT-3 and expert coders even when the model was given the codebook, and Tai et al.\ (\citeyear{tai2024examination}) reported similar gaps when applying GPT-3.5 to deductive coding. Other work uses LLMs to find codes that analysts might miss \cite{barany2024chatgpt}. These results point to the need for review by multiple coders, as human teams already do. 

\subsection{Peer Debriefing in QDA}

In qualitative research, a researcher's background shapes analytical outcomes. Braun and Clarke (\citeyear{braun2019reflecting}) argue that in reflexive thematic analysis, researcher subjectivity is not a threat but an analytical resource, as theoretical assumptions and analytic orientations actively shape coding and theme development. Reflexivity, the ongoing practice of treating one's background as an active part of the analysis rather than a bias to remove, is itself a methodological commitment in qualitative research \cite{olmos2023practical}. Empirical evidence supports this: Ortloff et al.\ (\citeyear{ortloff2023different}) found that researchers from different institutions produced substantively different codebooks from the same data, suggesting that disciplinary background, not just skill level, drives analytical variation.

In practice, this variation is managed through \textit{peer debriefing} \cite{lincoln1985naturalistic,hail2011peer}, a process in which colleagues from different disciplinary backgrounds, typically disinterested peers who are familiar with the research context but not invested in the analytical outcomes, review and challenge the coding from their respective standpoints. Recent HCI work points in the same direction: Reflexis \cite{ye2026reflexis} embeds reflexive prompts into collaborative coding so that disagreements between analysts surface as productive dialogues rather than errors to resolve. Our framework operationalizes this practice by introducing \peeragent that refines coding results from distinct analytical stances based on the self-reflection memos generated by the initial coding LLM, mirroring peer debriefers with different disciplinary orientations. 

\subsection{Multi-Agent Systems}

Recent advances in LLM-based multi-agent systems have demonstrated the value of multi-agent systems for complex tasks. Frameworks such as AutoGen \cite{wu2023autogen} and MetaGPT \cite{hong2023metagpt} enable structured multi-agent collaboration, while ChatEval \cite{chan2024chateval} use multi-agent argumentation for evaluation. For qualitative analysis, Thematic-LM \cite{qiao2025thematic} leverages multi-agent systems for theme generation. Advancing this line of research, our framework draws on qualitative research methodology, simulating the \textit{peer debriefing} process in its multi-agent design. Moreover, the perspectives assigned to our \peeragent are not domain-specific, making the framework broadly applicable across diverse datasets.


\section{Multi-Agent Framework}
\label{sec:framework}

\begin{figure*}[t]
  \centering
  \includegraphics[width=\textwidth]{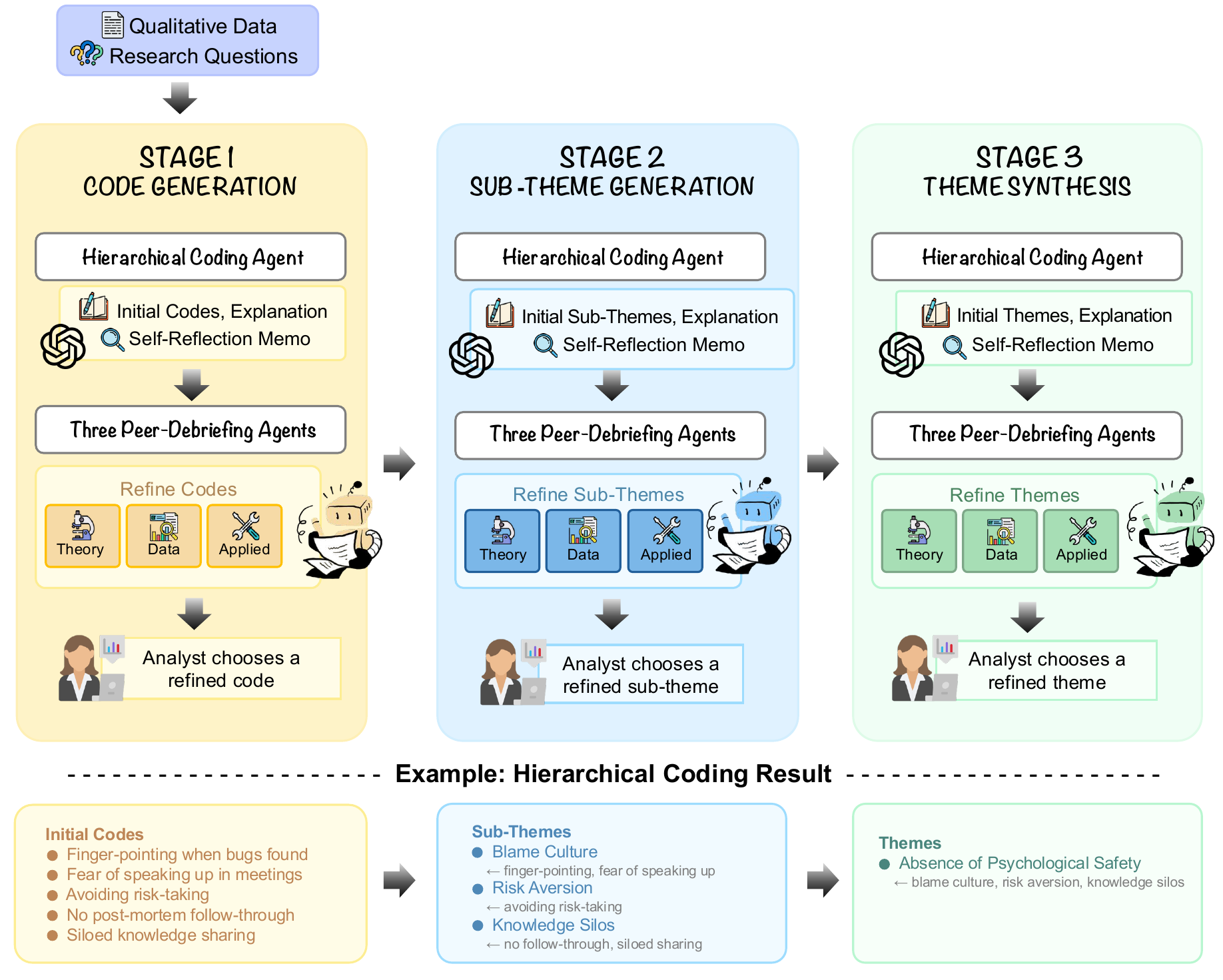}
  \caption{The framework consists of three hierarchical stages: code generation, sub-theme generation, and theme synthesis. At each stage, the \codingagent produces structured results with explanations and self-reflection memos. \peeragent{}s then re-examine and refine these results from different analytical perspectives. Researchers may stay in the loop to select the preferred refinement at each stage.}
  \label{fig:framework}
\end{figure*}

We propose a multi-agent framework for LLM-assisted QDA that incorporates \textit{peer debriefing} at every stage of the coding process. We first describe the overall framework (Section~\ref{sec:overview}), then the two agent types (Section~\ref{sec:agents}) and the analytical perspective modeling (Section~\ref{sec:perspectives}).

\subsection{Framework Overview}
\label{sec:overview}

Following the practice of \textit{peer debriefing} in qualitative research, where someone outside the original analysis reviews the coding, we add \peeragent{} to the standard LLM-assisted QDA setup. In this pipeline, a \codingagent generates initial codes, sub-themes, and themes, along with explanations and self-reflection memos. At each stage, these results are sent to three distinct \peeragent{}s, who review and refine the coding results based on these explanations and memos. 
It adopts three distinct analytical perspectives from established QDA practice: Theory-Driven, Data-Driven, or Applied (Section~\ref{sec:perspectives}). The framework runs all three in parallel at each stage, producing three refined coding structures. The human analysts may remain in the loop to select the most favorable one. Because these perspectives are domain-general, the framework transfers to new qualitative analysis tasks without rewriting the prompts; the only inputs that change across tasks are the research question and the interview data.

\subsection{Agent Design}
\label{sec:agents}

The framework has two types of agents: a \codingagent that produces the initial coding structure, and three \peeragent{}s that refine it from different analytical perspectives.

\paragraph{Hierarchical Coding Agent}
The \codingagent processes the qualitative data in three sequential stages. Aligning with the key steps of thematic analysis \cite{braun2006using,dai-etal-2023-llm, gao2025efficiencyrigortrustworthyllmpowered}, in the \textit{code generation} stage, it segments the interview text into analytical units and assigns each unit an initial code. In the \textit{sub-theme abstraction} stage, it groups related codes into sub-themes. In the \textit{theme synthesis} stage, it aggregates sub-themes into overarching themes. The same agent architecture runs at every stage with stage-specific instructions; only the input size and level of abstraction change. At each stage, the agent also writes explanations for codes it produces, together with self-reflection memo for the stage. These give the \peeragent the evidence and reasoning behind the coding decision. At each stage, the agent writes the self-reflection memo in a single pass, without further revising it; we leave iterative self-reflection to future work.

\paragraph{Peer-Debriefing Agent}
Each \peeragent takes on a specific analytical perspective. Given an existing coding structure and the \codingagent's reflection memo, it re-examines the relationships among codes, sub-themes, and themes from that perspective and may apply five operations:
\begin{itemize}[nosep]
    \item \textbf{Keep}: Retains the code without modification.
    \item \textbf{Rename}: Changes only the code name without altering its content or hierarchy.
    \item \textbf{Reassign}: Moves a code to a different parent within the current level.
    \item \textbf{Merge}: Combines multiple codes into a single code.
    \item \textbf{Split}: Divides one code into multiple codes.
\end{itemize}
These operations restructure the coding hierarchy without rewriting the original codes or the interview text. Instead of generating a new coding structure from scratch, the agent acts as a \textit{disinterested peer} who reinterprets the same structure from a specific perspective, so that differences between perspectives become directly comparable. We also do not include any human-generated codes as few-shot examples in the agent prompts; the prompts contain only general analytical instructions from established qualitative research methods, so the agents do not pick up any individual researcher's coding style. The full prompt templates are listed in Appendix~\ref{sec:prompt}.


\subsection{Perspective Modeling}
\label{sec:perspectives}
To keep the framework domain-general, we ground each perspective in a widely used qualitative research methodology \cite{saldana2009coding, corbin2017grounded} rather than tailoring it to a specific dataset or domain. We define three representative perspectives below. Each one applies a different \textit{evaluation criterion} when choosing candidate codes, and we illustrate them using the same interview excerpt:

\paragraph{Theory-Driven Perspective}
The theory-driven perspective \cite{braun2006using} is based on \textit{external knowledge}: theoretical frameworks and concepts from the relevant discipline. When two candidate codes conflict, the agent keeps the one with stronger theoretical grounding and better alignment with the literature. This mirrors researchers who read qualitative data through a theoretical lens. For example, given the interview statement \textit{``In our team, whenever a bug is found, people start pointing fingers at each other,''} the agent may produce a code such as \texttt{Absence of Psychological Safety}, which links the observation to the concept of psychological safety \cite{edmondson1999psychological}.

\paragraph{Data-Driven Perspective}
The data-driven perspective \cite{braun2006using} is based on \textit{local context}: the language and meaning in the data itself, without imposing any external framework. Codes emerge inductively from what participants say, and when two candidates conflict, the agent keeps the one that stays closer to the participant's wording. A related concept is ``in-vivo coding'', where researchers use participants' original words or phrases as codes \cite{saldana2009coding}. This mirrors researchers who do inductive qualitative analysis and stay close to the data. For the same interview excerpt, the agent may produce a code such as \texttt{Finger-Pointing When Bugs Are Found}, which preserves the phrasing without abstracting it into a theoretical label.

\paragraph{Applied Perspective}
The applied perspective emphasizes \textit{practical utility}: whether a code can directly inform a decision or an action. When candidates conflict, the agent keeps the one that more clearly suggests what a practitioner should do, even if another code is more theoretical or stays closer to the participant's wording. This mirrors practitioners who use qualitative analysis to inform decisions. The perspective is similar to ``Process code'' \cite{saldana2009coding}, which prioritizes actions and activities in the data. For the same excerpt, the agent may produce a code such as \texttt{Need for Blame-Free Post-Mortem Process}, which points to a concrete change a team lead could act on.

\begin{table}[t]
\setlength{\textfloatsep}{5pt}
\setlength{\abovecaptionskip}{4pt}
\setlength{\belowcaptionskip}{0pt}
\centering
\small
\setlength{\tabcolsep}{3pt}
\caption{Ground-truth annotation granularity of the evaluation datasets.}
\label{tab:dataset_gt}
\begin{tabular}{@{}lccc@{}}
\toprule
 & \textbf{ScrumRQ1} & \textbf{ScrumRQ2} & \textbf{UCSB} \\
\midrule
Domain & \multicolumn{2}{c}{Software Engineering} & Education \\
Interviews & \multicolumn{2}{c}{39} & 10 \\
Per-interview codes & \checkmark & \checkmark & -- \\
Overall Codebook & \checkmark & \checkmark & \checkmark \\
Overall Sub-themes & -- & -- & \checkmark \\
Overall Themes & -- & -- & \checkmark \\
\bottomrule
\end{tabular}
\end{table}

\section{Experiments}
\subsection{Experimental Settings}
\paragraph{Public Qualitative Datasets} We evaluate the framework on two public qualitative interview datasets from different domains.
The \textbf{Scrum} dataset\footnote{\url{https://zenodo.org/records/6624064}}~\cite{alami2022scrum} contains 39 semi-structured interviews with software professionals (Developers, Product Owners, Scrum Masters, and QA practitioners) on how Scrum contributes to software quality. We evaluate two research questions from the original study separately: ScrumRQ1 and ScrumRQ2. The dataset provides per-interview codes (each data chunk annotated with one or more codes) and a codebook with code definitions and frequency counts.
The \textbf{UCSB} dataset\footnote{\url{https://datadryad.org/dataset/doi:10.25349/D9402J}}~\cite{curty2021teaching} contains 10 semi-structured interviews with university instructors on quantitative data teaching in the social sciences. It provides a codebook of codes organized into sub-themes and themes, which allows evaluation at the thematic level. Table~\ref{tab:dataset_gt} summarizes the two datasets. Together, they let us evaluate coding quality at the per-interview level (Scrum) and thematic coherence at the sub-theme and theme level (UCSB), while codebook-level metrics provide cross-dataset comparability.

To our knowledge, this is one of the most comprehensive evaluation setups in the LLM-assisted qualitative analysis literature to date. Most existing studies evaluate on a single dataset \cite{piano-etal-2023-qualitative, retkowski-etal-2025-ai} at the code level only, often using short text segments such as survey responses or quote-code pairs rather than full interview transcripts \cite{parfenova-etal-2025-text, parfenova-pfeffer-2025-measuring}. In contrast, our evaluation spans three datasets from two domains and covers all three levels of thematic analysis (code, sub-theme, and theme) across three LLM backbones. On Scrum, we also generate codes per interview and compare them against the ground truth. All three datasets include published human-annotated ground-truth codes, which allows direct quantitative comparison at each analytical level and makes our results reproducible.

\begin{figure*}[t]
\centering
\begin{tcolorbox}[
    enhanced,
    title=Interview Excerpt Example (Scrum Dataset),
    fonttitle=\bfseries\small,
    colback=white,
    colframe=black,
    boxrule=0.5pt,
    arc=2pt,
    left=4pt,
    right=4pt,
    top=4pt,
    bottom=4pt
]
\small
To illustrate the nature of the qualitative data used in this work, we provide a brief excerpt from a semi-structured interview in the Scrum dataset. The following is a sample exchange between a researcher (R) and a software practitioner (P) on the topic of software quality and Scrum practices:
\tcblower
\small
\begin{Verbatim}[breaklines=true, fontsize=\small]
R: What's the difficulties of implementing Agile, which is a culture, actually? What are the difficulties of implementing such a culture in companies that sometimes are difficult to change?

P: I think mainly getting everyone on board with what you want to do is probably the biggest hurdle. It also depends on what it is exactly you want to do. Agile is quickly used as a word, as a Scrum, as a Scrumban. It's differently understood by different people. And usually management wants one thing, the developers want another thing. They both have very different opinions about what is what. And we try to find the middle road or the best road towards delivering a product to the customers.
\end{Verbatim}
\end{tcolorbox}
\caption{A sample exchange between a researcher (R) and a software practitioner (P) from a semi-structured interview in the Scrum dataset, illustrating the qualitative data used in this work.}
\label{fig:interview_excerpt}
\end{figure*}

\paragraph{Base Models} We run the framework with three LLMs as the agent backbone: GPT-5 \cite{singh2025openai}, DeepSeek-V3.2 \cite{deepseekai2025deepseekv32}, and Qwen3.5-Plus \cite{qwen35blog}. This lets us compare how different base models affect coding quality under the same framework.

\paragraph{Perspective Isolation} We evaluate each perspective independently to isolate its effect on coding quality. As baselines, we compare against single-pass coding with no refinement and a refinement pass without any perspective. In practice, researchers can mix perspectives across stages (e.g., choosing a theory-driven result at the code level and an applied result at the sub-theme level). We leave mixed-perspective strategies to future work.

\subsection{Metrics Definition}
We evaluate the quality of generated codes with three metrics drawn from prior work~\cite{gao2023coaicoder, gebreegziabher2023patat}: Recall, Match Rate, and Code Diversity. Throughout this section, let $G = \{g_1, g_2, \dots\}$ denote the set of codes the model generates and $H = \{h_1, h_2, \dots\}$ the set of human-annotated codes (the ground truth). Recall and Match Rate compare $G$ against $H$ through a code-level matching procedure described next; Code Diversity is computed on $G$ alone.

\paragraph{Code-level Matching} Intuitively, for each generated code we look up its closest human code by meaning, and treat the two as the same code if they are similar enough. Formally, for each generated code $g_i \in G$, we compute its cosine similarity with every human-annotated code $h_j \in H$, using all-MiniLM-L6-v2 \cite{wang2020minilm} as the text encoder. We normalize each score to $[0,1]$ via $\tilde{s}_{i,j} = (s_{i,j}+1)/2$ and take the best match $j^* = \arg\max_{j} \tilde{s}_{i,j}$. We then define a matching function $M : G \to H \cup \{\varnothing\}$: if $\tilde{s}_{i,j^*} \ge \tau$ (we set $\tau = 0.7$), $M(g_i) = h_{j^*}$; otherwise $M(g_i) = \varnothing$, meaning $g_i$ has no match. The matching is one-to-many: each generated code matches at most one human code, but a single human code can be matched by several generated codes. We chose $\tau = 0.7$ to be permissive enough for paraphrase but strict enough to exclude pairs that are topically related but conceptually distinct, and verified it by manually inspecting boundary pairs (similarity in $0.65$--$0.75$).

\paragraph{Recall}
Intuitively, Recall asks: of all the codes a human produced ($|H|$, fixed in advance), how many did the model recover? Formally, Recall is the fraction of human-annotated codes matched by at least one generated code:
\[
\text{Recall} = \frac{|\{j : \exists\, i,\ M(g_i) = h_j\}|}{|H|},
\]
where $g_i$ is a generated code, $h_j$ a human-annotated code, and $M(\cdot)$ the matching function defined above (so $M(g_i) = h_j$ means $g_i$ was matched to $h_j$).

\paragraph{Match Rate} Intuitively, Match Rate asks: of all the codes the model itself produced ($|G|$, which the model controls), how many align with a code in the human codebook? 
Formally, it is the fraction of generated codes that match some human code:
\[
\text{Match Rate} = \frac{|\{i : M(g_i) \neq \varnothing\}|}{|G|},
\]
where $M(g_i) \neq \varnothing$ means generated code $g_i$ found a human-annotated code with similarity above $\tau$.

We use the term Match Rate rather than precision because all generated codes are grounded in the interview text and thus analytically valid; a generated code that does not match any human code is not necessarily a false positive but may reflect a valid interpretation not captured by the human codebook. Match Rate therefore measures the degree of alignment with the human reference rather than correctness per se.

\paragraph{Code Diversity} Intuitively, Code Diversity asks: of all the codes the model produced, how many are semantically distinct from the others? We iteratively remove one code from each pair of generated codes whose similarity $\tilde{s}(g_i, g_k) \ge \tau$, yielding a deduplicated set $G^{*}$:
\[
\text{Code Diversity} = \frac{|G^{*}|}{|G|},
\]
where $G$ is the full set of generated codes and $G^{*}$ is its deduplicated subset (with near-duplicate pairs collapsed to one). Higher values indicate greater independence among generated codes.

\subsection{Results}
We report our main results in two parts, corresponding to our two research questions (RQ1 and RQ2). For each RQ, we compare the three perspectives (Theory-Driven, Data-Driven, and Applied) against a no-refinement baseline across all models and datasets. We additionally run a Self-Refinement ablation on GPT-5 (same prompt structure as the \peeragent but without any perspective framing; see \S\ref{sec:ablation}) to test whether gains stem from perspective or from a second refinement pass. We analyze the results using the three evaluation metrics defined above. All reported means are averaged over 39 interviews for the Scrum RQ1 and Scrum RQ2 datasets, and 10 interviews for UCSB.

\begin{figure*}[!htbp]
  \centering
  \includegraphics[width=1.85\columnwidth]{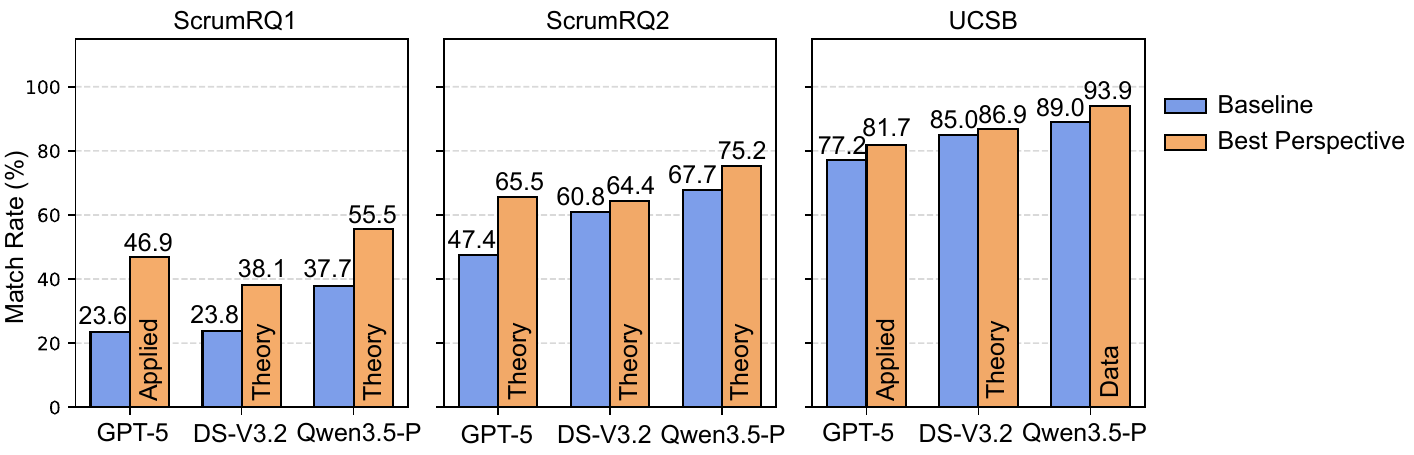}
  \caption{Match Rate comparison between baseline (no perspective) and the best-performing perspective for each model--dataset combination. The best-performing perspective for each configuration is identified from the full results reported in Table~\ref{tab:match_rate}. DS-V3.2 denotes DeepSeek-V3.2; Qwen3.5-P denotes Qwen3.5-Plus.}
  \label{fig:rq1}
\end{figure*}

\subsubsection{RQ1: Does peer-debriefing improve AI coding quality?}
We use Match Rate to answer RQ1, as it means the proportion of AI-generated codes that have similar words to the human codebook.
Our results show that our approach is promising to improve the Match Rate over the baseline at the code level (Appendix Table~\ref{sec:experimental_results}). In particular, the best-performing perspective can substantially improve Match Rate over the baseline (Figure~\ref{fig:rq1}). \textbf{This means that of all AI-generated codes, a greater proportion have similar words to the human codebook.}
For example, on the ScrumRQ1 dataset, the Match Rate of GPT-5 increases from 23.6\% to 46.9\% (Applied), while Qwen3.5-Plus improves Match Rate from 37.7\% to 55.5\% (Theory-Driven). On the ScrumRQ2 dataset, GPT-5 further increases Match Rate from 47.4\% to 65.5\% (Theory-Driven). For other ones that are not best-performing, they remain aligned with the baseline or show only slight gains over the baseline. For example, the Data-Driven perspective leaves Qwen3.5-Plus's Match Rate on ScrumRQ1 unchanged (37.7\%$\to$37.7\%), and the Applied perspective raises DeepSeek-V3.2's Match Rate on UCSB by less than a point (85.0\%$\to$85.1\%). The same Applied perspective raises GPT-5's Match Rate on UCSB by more than four points (77.2\%$\to$81.7\%).

Overall, on Match Rate, the approach is promising in improving AI coding quality. Other dimensions, Recall and Code Diversity could be more informative on measuring different aspects of the refinement capability. Next, we show the differences among different perspectives.


\subsubsection{RQ2: How do different analytical perspectives affect coding outcomes?}


\paragraph{The Theory-Driven and Applied perspectives generated more matched codes with humans than baseline, but fewer codes in the human codebook can be recalled.} On the code level, the two perspectives made changes to AI-generated codes, and these changes, as we observed, made those codes more aligned with what humans would write. Due to their merging, renaming, etc. operations, these codes were more abstract toward higher-level interpretations, which helps align with human coding decisions. But these operations also result in fewer codes from the human codebook being recalled. As shown in the data, these two perspectives consistently produce the highest Match Rates at the code level (e.g., GPT-5 on ScrumRQ1: Theory 44.4\%, Applied 46.9\%, vs.\ baseline 23.6\%). 


\paragraph{Data-Driven aligns more with raw data at the code level, but has more themes extracted from the original human codebook.} The Data-Driven perspective was designed to keep alignment with the original data's meaning. At the code level, its codes are more grounded in participants' language, so that the agent tends to \textit{Keep} instead of changing the code (as shown in Table \ref{tab:case_study}). This thus leads to a modest Match Rate at the code level (28.2\% vs. 23.6\% baseline), implying that many codes do not directly correspond to the human codebook. But this causes higher Recall at the code level. And thus it also causes interesting Recall at higher-level themes.
At the theme level, due to the Data-Driven approach generating themes on more diverse codes from their prior stage, its generated and refined themes have the largest recall gains (GPT-5: 40.0\%$\to$60.0\%; DS-V3.2: 60.0\%$\to$70.0\%) and Match Rates (DS-V3.2: 96.4\%).

\paragraph{The effect of all perspectives grows stronger at higher abstraction levels.} Across all three perspectives, refinement seems to have a stronger effect at higher abstraction levels. At the theme level, Recall improvements reach +20 percentage points (GPT-5). This suggests that perspective-driven refinement is particularly effective where analytical interpretation is highly required.

\subsection{Preliminary Ablation Study}
\label{sec:ablation}
To disentangle the effect of analytical perspective from an additional refinement pass, we conduct an ablation using a \textbf{Self-Refinement} baseline. This baseline uses the same prompt structure as the \peeragent (including all five structural operations and the self-reflection memo as input) but removes all perspective-specific framing. If the improvements reported in RQ1 were simply due to ``refining one more time,'' Self-Refinement should perform comparably to the perspective conditions.

We run this ablation with GPT-5 across three datasets at the code level (Table~\ref{tab:self-refine} in Appendix). Self-Refinement does improve over the baseline, confirming that an additional refinement pass has value (e.g., Match Rate on ScrumRQ1: 23.6\%$\to$32.3\%; ScrumRQ2: 47.4\%$\to$59.0\%). However, perspective-based refinement consistently outperforms Self-Refinement: on ScrumRQ1, Theory (44.4\%) and Applied (46.9\%) exceed Self-Refinement (32.3\%) by a substantial margin. On ScrumRQ2, Theory (65.5\%) and Applied (64.2\%) similarly surpass Self-Refinement (59.0\%). These results indicate that the gains from perspective-based refinement cannot be attributed solely to self-correction; the explicit modeling of analytical perspectives provides value beyond what an undirected refinement pass achieves.

\subsection{Qualitative Case Study}
To show how perspective-based refinement works in practice, we walk through one example from Scrum RQ1 (Table~\ref{tab:case_study}). Starting from the same 10 baseline codes, all three perspectives independently flag the \texttt{Retrospectives} code as problematic because it mixes participant statements with researcher questions, and each debriefing agent splits it to separate the two voices. 

We summarized how the three perspectives diverge in Table \ref{tab:mechanism}. In particular, the perspectives diverge in how they consolidate and rename. The Theory perspective merges testing-related codes into one higher-level concept; the Applied perspective goes further and relabels codes in practitioner-oriented language. The baseline code \texttt{Process vs product} is renamed differently by each perspective, mirroring how human researchers from different backgrounds would label the same idea.

\begin{table}[H]
\centering
\footnotesize
\setlength{\tabcolsep}{5pt}
\caption{How the three perspectives diverge. Refer to Figure \ref{fig:heatmap_ops} and Table \ref{tab:operation-breakdown} for more details.}
\label{tab:mechanism}
\begin{tabular}{@{}lrrl@{}}
\toprule
 & \multicolumn{2}{c}{\shortstack{\textbf{Operations}\\\textbf{per interview}}} & \\
\cmidrule(lr){2-3}
\textbf{Perspective} & \textbf{Merge} & \textbf{Split} & \textbf{Renames toward} \\
\midrule
Theory-Driven & 1.48 & 0.46 & theoretical concepts \\
Applied       & 1.39 & 0.50 & practitioner vocabulary \\
Data-Driven   & \textbf{1.00} & \textbf{0.70} & raw data language \\
\bottomrule
\end{tabular}
\end{table}

\begin{table*}[t]
\centering
\small
\setlength{\tabcolsep}{5pt}
\renewcommand{\arraystretch}{1.15}

\caption{An example of perspective-driven code refinement on Scrum Interview.}
\label{tab:case_study}

\begin{tabularx}{\textwidth}{@{}
  >{\raggedright\arraybackslash}p{0.17\textwidth}
  *{3}{>{\raggedright\arraybackslash}X}@{}}
\toprule
\textbf{Baseline code(s)} & \textbf{Theory-driven} & \textbf{Applied} & \textbf{Data-driven} \\
\midrule

Process vs.\ product
  & \oprename Quality dimensions: process, product, and internal code quality
  & \oprename Process- and product-level quality definition
  & \oprename Process and product dimensions of quality \\
\addlinespace[5pt]

Envs and unit tests\newline{}+ Canary beta releases
  & \opmerge Staged testing and release process
  & \opmerge Multi-environment testing, unit tests, and staged releases
  & \opkeep \\
\addlinespace[5pt]

Transparency view\newline{}+ Naive Scrum harms
  & \opkeep
  & \opmerge Agile misconceptions and transparency shaping perceived quality
  & \opmerge Agile misuse and transparency shaping perceived quality outcomes \\
\addlinespace[5pt]

Retrospectives
  & \opsplit Continuous improvement via retrospectives\newline
    \sub{interviewer framing on quality enablers}
  & \opsplit Continuous improvement via retrospectives\newline
    \sub{researcher framing: context, not practitioner-defined}
  & \opsplit Retrospectives for continuous improvement\newline
    \sub{interviewer framing on quality gaps and enablers} \\
\bottomrule
\end{tabularx}
\end{table*}

\section{Discussion}
\label{sec:discussion}

To see how different analytical perspectives shape coding, we look at the edits the \peeragent makes to the code structure. As described in Section~\ref{sec:framework}, the agent has five operations: \textit{Keep}, \textit{Rename}, \textit{Reassign}, \textit{Merge}, and \textit{Split}. \textit{Keep} leaves the structure as is, so we focus on the four operations that change it. Table \ref{tab:operation-breakdown} and Figure~\ref{fig:heatmap_ops} report more detailed operations across models, perspectives, and datasets.

\paragraph{Rename}
\textit{Rename} is the most frequent operation. In most settings, the agent uses \textit{Rename} more often than \textit{Merge} or \textit{Split}. For example, in ScrumRQ1, GPT-5 (Applied) does 4.6 \textit{Renames} on average, compared with 2.3 \textit{Merges} and 1.0 \textit{Split}.
The agent therefore tends to adjust the words attached to a code rather than reshape the code structure. This matches how human qualitative researchers usually work: they revise code labels repeatedly to better fit the data.

\paragraph{Reassign} \textit{Merge} and \textit{Split} rearrange the code structure itself. \textit{Reassign} is different: it moves a text chunk to another code, or moves a code under a different sub-theme or theme, without changing the codes themselves. We count \textit{Reassign} only when it happens on its own, not when \textit{Merge} or \textit{Split} implicitly moves chunks as a side effect. How often \textit{Reassign} is used varies a lot across models. GPT-5 uses it most heavily (1.20--2.33 per interview), making many small adjustments, while DeepSeek-V3.2 (0.67--1.02) and Qwen3.5-Plus (0.00--0.60) lean on \textit{Merge} instead.

\paragraph{Merge} \textit{Merge} happens less often than \textit{Rename}, but it is more closely tied to theory-oriented perspectives. For example, in ScrumRQ1, DeepSeek-V3.2 (Theory) does 1.9 \textit{Merges} on average, more than under other perspectives.
\textit{Merging} groups related codes into a higher-level category, which is what theory-driven analysis tends to do: pulling specific observations together into broader concepts.

\paragraph{Split}
\textit{Split} is the rarest operation in most settings, usually fewer than one per interview. A few perspective-dataset pairs push higher: on UCSB, GPT-5 (Data) does 2.2 \textit{Splits} on average.
\textit{Splitting} breaks one code into more specific subcodes. When a baseline code lumps together several different things, the data-driven perspective tends to pull them apart.

\section{Conclusion and Future Work}

To bring LLM-generated codes closer to human coding, we adapt the qualitative research practice of \textit{peer debriefing}, where researchers seek feedback from several disinterested peers to gain feedback and refine their own codes. We then propose \systemname, a multi-agent framework that simulates it in LLM-assisted QDA. Across three datasets and three base models, perspective-based refinement consistently improves AI coding quality. Moreover, different perspectives yield distinct trade-offs, making the choice of perspective a meaningful and controllable design decision. Future work could let researchers consult peer agents interactively, incorporate human feedback into the refinement loop, or aggregate refinements through a multi-agent debate, where peer agents collectively deliberate on code modifications.

\section*{Limitations}
While our framework demonstrates the value of perspective-based refinement, it also has limitations. First, the framework relies on the quality of the initial coding structure generated by the Hierarchical Coding Agent; different prompt designs may lead to different initial coding structures, which may influence the subsequent refinement process. Second, while we analyze the frequency of revision operations, there is an opportunity for future work to incorporate human evaluation to assess how perspective-based revisions affect the meaningfulness and utility of the final coding scheme.

\bibliography{custom}
\appendix
\onecolumn

\clearpage
\section{Experimental Results}
\label{sec:experimental_results}

\subsection{Code-Level Comparison}
We report four code-level metrics (recall, match rate, code diversity, and text utilization rate) below.

\subsubsection{Recall}
Perspective-guided refinement, especially under the Data-driven perspective, tends to match or improve recall relative to the baseline across most model-dataset combinations (Table~\ref{tab:recall}).
\begin{center}
\begin{minipage}{\textwidth}
\centering
\small
\setlength{\tabcolsep}{5pt}
\captionof{table}{
Recall at the code level for baseline and perspective-refined codebooks on the ScrumRQ1, ScrumRQ2, and UCSB datasets.
}
\label{tab:recall}
\begin{tabular}{lccc}
\toprule
\textbf{Model / Perspective} & \textbf{ScrumRQ1} & \textbf{ScrumRQ2} & \textbf{UCSB} \\
\midrule
GPT-5 & 70.6 & 80.0 & 42.3 \\
GPT-5 (Theory) & 82.4 & 80.0 & 35.9 \\
GPT-5 (Applied) & 76.5 & 80.0 & 38.5 \\
GPT-5 (Data) & \textbf{88.2} & \textbf{85.0} & \textbf{47.4} \\
\midrule
DS-V3.2 & 82.4 & \textbf{77.5} & 32.1 \\
DS-V3.2 (Theory) & 76.5 & \textbf{77.5} & 24.4 \\
DS-V3.2 (Applied) & 76.5 & 70.0 & 28.2 \\
DS-V3.2 (Data) & \textbf{88.2} & 72.5 & \textbf{33.3} \\
\midrule
Qwen3.5-P & 70.6 & \textbf{87.5} & 35.9 \\
Qwen3.5-P (Theory) & 58.8 & 82.5 & 25.6 \\
Qwen3.5-P (Applied) & \textbf{76.5} & 82.5 & 30.8 \\
Qwen3.5-P (Data) & \textbf{76.5} & \textbf{87.5} & \textbf{37.2} \\
\bottomrule
\end{tabular}
\end{minipage}
\end{center}

\subsubsection{Match Rate}
Match Rate increases in almost all settings across models and perspectives. Across all three backbone models, the Theory and Applied perspectives tend to raise match rate the most.
\begin{center}
\begin{minipage}{\textwidth}
\centering
\small
\setlength{\tabcolsep}{5pt}
\captionof{table}{
Match rate at the code level for base and perspective-refined codebooks.
}
\label{tab:match_rate}
\begin{tabular}{lccc}
\toprule
\textbf{Model / Perspective} & \textbf{ScrumRQ1} & \textbf{ScrumRQ2} & \textbf{UCSB} \\
\midrule
GPT-5 & 23.6 & 47.4 & 77.2 \\
GPT-5 (Theory) & 44.4 & \textbf{65.5} & 77.8 \\
GPT-5 (Applied) & \textbf{46.9} & 64.2 & \textbf{81.7} \\
GPT-5 (Data) & 28.2 & 53.2 & 77.7 \\
\midrule
DS-V3.2 & 23.8 & 60.8 & 85.0 \\
DS-V3.2 (Theory) & \textbf{38.1} & \textbf{64.4} & \textbf{86.9} \\
DS-V3.2 (Applied) & 34.3 & 62.8 & 85.1 \\
DS-V3.2 (Data) & 27.3 & 62.5 & 86.5 \\
\midrule
Qwen3.5-P & 37.7 & 67.7 & 89.0 \\
Qwen3.5-P (Theory) & \textbf{55.5} & \textbf{75.2} & 89.6 \\
Qwen3.5-P (Applied) & 46.8 & 72.5 & 90.2 \\
Qwen3.5-P (Data) & 37.7 & 68.1 & \textbf{93.9} \\
\bottomrule
\end{tabular}
\end{minipage}
\end{center}

\subsubsection{Code Diversity}
The unrevised baseline generally retains slightly higher code diversity than perspective-refined codebooks, indicating that refinement consolidates codes into more focused categories (Table~\ref{tab:code_diversity}).
\begin{center}
\begin{minipage}{\textwidth}
\centering
\small
\setlength{\tabcolsep}{5pt}
\captionof{table}{
Code diversity of base and perspective-revised codebooks at the code level.
}
\label{tab:code_diversity}
\begin{tabular}{lccc}
\toprule
\textbf{Model / Perspective} & \textbf{ScrumRQ1} & \textbf{ScrumRQ2} & \textbf{UCSB}  \\
\midrule
GPT-5 & \textbf{25.9} & \textbf{22.1} & \textbf{34.7} \\
GPT-5 (Theory) & 14.3 & 11.6 & 25.2 \\
GPT-5 (Applied) & 11.2 & 12.1 & 23.0 \\
GPT-5 (Data) & 18.0 & 18.3 & 34.2 \\
\midrule
DS-V3.2 & \textbf{16.9} & 12.3 & 19.0 \\
DS-V3.2 (Theory) & 12.0 & 12.0 & 19.0 \\
DS-V3.2 (Applied) & 14.6 & 12.8 & \textbf{24.1} \\
DS-V3.2 (Data) & 14.1 & \textbf{12.9} & 20.0 \\
\midrule
Qwen3.5-P & 14.9 & 12.6 & 21.0 \\
Qwen3.5-P (Theory) & 12.6 & 12.5 & \textbf{22.7} \\
Qwen3.5-P (Applied) & 13.1 & 11.8 & 19.6 \\
Qwen3.5-P (Data) & \textbf{15.0} & \textbf{13.6} & 22.4 \\
\bottomrule
\end{tabular}
\end{minipage}
\end{center}

\subsubsection{Code Diversity}
Baseline coding without perspective guidance generally maintains slightly higher code diversity (Table~\ref{tab:code_diversity}), implying that perspective-based refinement leads the model toward more focused, consolidated coding patterns. To verify that this pattern is not an artifact of the order in which duplicate codes are iteratively removed, we repeat the deduplication procedure for GPT-5 across all 12 configurations under five random seeds (Table~\ref{tab:diversity_robustness}); the resulting standard deviations are small (at most 1.9 points), confirming that the reported Code Diversity values are robust to this ordering choice.

\begin{table*}[!h]
\centering
\small
\setlength{\tabcolsep}{5pt}
\caption{Code Diversity robustness analysis of GPT-5 across all 12 configurations (4 conditions $\times$ 3 datasets) under 5 fixed random seeds. Each seed is used to shuffle the order of generated codes prior to the iterative removal procedure. Values are reported as percentages (\%). The maximum standard deviation is 1.9, confirming that the removal order has negligible impact on the reported Code Diversity values.}
\label{tab:diversity_robustness}
\begin{tabular}{lcccccc}
\toprule
\textbf{Configuration} & \textbf{Seed 0} & \textbf{Seed 42} & \textbf{Seed 777} & \textbf{Seed 2026} & \textbf{Seed 99999} & \textbf{Mean $\pm$ Std} \\
\midrule
ScrumRQ1 (Initial)  & \textbf{26.4} & \textbf{25.9} & \textbf{25.4} & \textbf{24.9} & \textbf{25.1} & 25.5 $\pm$ 0.6 \\
ScrumRQ1 (Applied)  & 13.3 & 13.6 & 14.5 & 12.3 & 13.3 & 13.4 $\pm$ 0.7 \\
ScrumRQ1 (Theory)  & 12.5 & 12.8 & 11.6 & 13.1 & 12.2 & 12.4 $\pm$ 0.5 \\
ScrumRQ1 (Data)  & 18.5 & 22.1 & 18.7 & 20.8 & 17.9 & 19.6 $\pm$ 1.6 \\
\midrule
ScrumRQ2 (Initial)  & \textbf{22.3} & \textbf{22.6} & \textbf{22.8} & \textbf{22.6} & \textbf{21.8} & 22.4 $\pm$ 0.3 \\
ScrumRQ2 (Applied)  & 13.6 & 12.3 & 14.4 & 13.4 & 13.6 & 13.5 $\pm$ 0.6 \\
ScrumRQ2 (Theory)  & 11.8 & 11.6 & 12.6 & 11.3 & 10.6 & 11.6 $\pm$ 0.7 \\
ScrumRQ2 (Data)  & 17.8 & 16.2 & 17.1 & 17.6 & 17.1 & 17.2 $\pm$ 0.5 \\
\midrule
UCSB (Initial)  & \textbf{33.7} & \textbf{36.6} & \textbf{39.6} & \textbf{35.6} & \textbf{35.6} & 36.2 $\pm$ 1.9 \\
UCSB (Applied)  & 26.2 & 27.0 & 27.0 & 27.8 & 24.6 & 26.5 $\pm$ 1.1 \\
UCSB (Theory)  & 22.0 & 22.0 & 23.6 & 22.0 & 24.4 & 22.8 $\pm$ 1.0 \\
UCSB (Data)  & 31.5 & 30.8 & 32.9 & 30.1 & 34.2 & 31.9 $\pm$ 1.5 \\

\bottomrule
\end{tabular}
\end{table*}

\subsubsection{Text Utilization Rate}
Text utilization rate remains largely stable across perspectives within each backbone model, suggesting that it is affected primarily by the choice of base model and dataset.
\begin{center}
\begin{minipage}{\textwidth}
\centering
\small
\setlength{\tabcolsep}{5pt}
\captionof{table}{
Text utilization rate at the code level for base and perspective-revised codebooks on the ScrumRQ1, ScrumRQ2, and UCSB datasets.
}
\label{tab:text_utilization_rate}
\begin{tabular}{lccc}
\toprule
\textbf{Model / Perspective} & \textbf{ScrumRQ1} & \textbf{ScrumRQ2} & \textbf{UCSB}  \\
\midrule
GPT-5 & \textbf{74.2} & 75.6 & \textbf{87.6} \\
GPT-5 (Theory) & 74.1 & \textbf{77.3} & \textbf{87.6} \\
GPT-5 (Applied) & \textbf{74.2} & 74.7 & 87.4 \\
GPT-5 (Data) & 74.0 & 75.6 & 87.5 \\
\midrule
DS-V3.2 & 44.8 & \textbf{55.4} & \textbf{39.1} \\
DS-V3.2 (Theory) & 42.8 & \textbf{55.4} & \textbf{39.1} \\
DS-V3.2 (Applied) & \textbf{47.6} & 55.1 & 29.3 \\
DS-V3.2 (Data) & 44.8 & 55.3 & \textbf{39.1} \\
\midrule
Qwen3.5-P & 22.7 & \textbf{33.1} & \textbf{23.3} \\
Qwen3.5-P (Theory) & \textbf{23.0} & 32.9 & \textbf{23.3} \\
Qwen3.5-P (Applied) & 21.1 & 33.0 & 23.2 \\
Qwen3.5-P (Data) & 22.5 & 32.8 & 23.2 \\
\bottomrule
\end{tabular}
\end{minipage}
\end{center}

\subsection{Sub-Theme-Level Comparison}
At the sub-theme level (Table~\ref{tab:subtheme_compare}), the benefit of perspective-guided refinement becomes more model-dependent: the Data perspective still yields the largest gains in recall and code diversity for GPT-5, while for DeepSeek-V3.2 and Qwen3.5-Plus the strongest perspective shifts to Theory and Applied, respectively, particularly for match rate. Text utilization rate again changes little across perspectives within a given backbone model, consistent with the code-level results above.

\begin{center}
\begin{minipage}{\textwidth}
\centering
\small
\setlength{\tabcolsep}{5pt}

\captionof{table}{
Performance comparison on the UCSB dataset at the sub-theme level, between the original codebooks generated by three base models and those revised by \peeragent.
Text Utilization Rate measures the coverage of the original corpus during coding. We collect all deduplicated text segments referenced by generated codes into $T_{used}$, and compute $\text{Text Utilization Rate} = |T_{used}| / |T_{all}|$, where $|T_{used}|$ and $|T_{all}|$ denote the token counts of the used and full corpus, respectively.
}

\label{tab:subtheme_compare}

\begin{tabular}{lcccc}
\toprule
\textbf{Model / Perspective} & \textbf{Recall} & \textbf{Match Rate} & \textbf{Code Diversity} & \textbf{Text Utilization Rate} \\
\midrule
GPT-5 & 55.6 & \textbf{70.4} & 32.4 & \textbf{87.6} \\
GPT-5 (Theory) & 55.6 & 68.5 & 32.9 & \textbf{87.6} \\
GPT-5 (Applied) & 55.6 & 66.2 & 28.4 & 87.4 \\
GPT-5 (Data) & \textbf{61.1} & 68.6 & \textbf{34.3} & 87.5 \\
\midrule
DS-V3.2 & \textbf{55.6} & 82.1 & 21.4 & \textbf{39.1} \\
DS-V3.2 (Theory) & 44.4 & \textbf{86.7} & 20.0 & \textbf{39.1} \\
DS-V3.2 (Applied) & 50.0 & 84.1 & \textbf{22.7} & 29.3 \\
DS-V3.2 (Data) & 50.0 & 80.0 & 20.0 & \textbf{39.1} \\
\midrule
Qwen3.5-P & 50.0 & 94.2 & 21.2 & \textbf{23.3} \\
Qwen3.5-P (Theory) & 50.0 & 91.1 & 15.6 & \textbf{23.3} \\
Qwen3.5-P (Applied) & \textbf{61.1} & \textbf{95.2} & 21.4 & 23.2 \\
Qwen3.5-P (Data) & 55.6 & 83.3 & \textbf{25.0} & 23.2 \\
\bottomrule
\end{tabular}

\end{minipage}
\end{center}

\clearpage

\subsection{Theme-Level Comparison}
At the coarser theme level (Table~\ref{tab:theme_compare}), perspective-guided refinement matches or improves recall over the unrevised baseline for most settings, most consistently under the Theory and Data perspectives. Match rate shows a similarly mixed pattern to the sub-theme level, with the best-performing perspective again varying across models (baseline for GPT-5, Data for DeepSeek-V3.2, and Applied for Qwen3.5-Plus), while code diversity and text utilization rate remain comparatively stable across perspectives.

\begin{table*}[!h]
\centering
\small
\setlength{\tabcolsep}{5pt}
\caption{Performance comparison on the UCSB dataset at the theme level, between the original codebooks generated by three base models and the codebooks revised by \peeragent.}
\label{tab:theme_compare}
\begin{tabular}{lcccc}
\toprule
\textbf{Model / Perspective} & \textbf{Recall} & \textbf{Match Rate} & \textbf{Code Diversity} & \textbf{Text Utilization Rate} \\
\midrule
GPT-5 & 40.0 & \textbf{80.0} & \textbf{23.3} & \textbf{87.6} \\
GPT-5 (Theory) & 40.0 & 64.5 & 19.4 & \textbf{87.6} \\
GPT-5 (Applied) & 50.0 & 76.7 & 16.7 & 87.4 \\
GPT-5 (Data) & \textbf{60.0} & 78.1 & 21.9 & 87.5 \\
\midrule
DS-V3.2 & 60.0 & 86.7 & 16.7 & \textbf{39.1} \\
DS-V3.2 (Theory) & \textbf{70.0} & 80.0 & \textbf{20.0} & \textbf{39.1} \\
DS-V3.2 (Applied) & 50.0 & 93.1 & 10.3 & 29.3 \\
DS-V3.2 (Data) & \textbf{70.0} & \textbf{96.4} & 17.9 & \textbf{39.1} \\
\midrule
Qwen3.5-P & \textbf{60.0} & 83.3 & 20.0 & \textbf{23.3} \\
Qwen3.5-P (Theory) & \textbf{60.0} & 68.0 & \textbf{24.0} & \textbf{23.3} \\
Qwen3.5-P (Applied) & 40.0 & \textbf{89.3} & 17.9 & 23.2 \\
Qwen3.5-P (Data) & \textbf{60.0} & 86.2 & 20.7 & 23.2 \\
\bottomrule
\end{tabular}
\end{table*}

\subsection{Comparison with Self-Refinement}
Table~\ref{tab:self-refine} isolates the contribution of external perspectives from generic self-reflection by comparing GPT-5's baseline codebook against a Self-Refine variant (self-reflection without additional perspectives). Self-Refine alone yields at most marginal gains over the baseline and does not consistently outperform it on any metric, whereas perspective-guided refinement, particularly Theory for match rate and Data for recall, delivers larger and more consistent improvements. This suggests that the gains from \peeragent stem from the external perspectives themselves rather than from self-reflection alone.

\begin{table*}[!h]
\centering
\small
\setlength{\tabcolsep}{5pt}
\caption{Performance comparison of GPT-5 under different refinement strategies across three datasets (ScrumRQ1, ScrumRQ2, and UCSB).
"Self-Refine" denotes self-reflection without additional perspectives, while Theory, Applied, and Data represent perspective-guided refinement.}
\label{tab:self-refine}
\begin{tabular}{llcccc}
\toprule
\textbf{Dataset} & \textbf{Model / Perspective} & \textbf{Recall} & \textbf{Match Rate} & \textbf{Code Diversity} & \textbf{Text Utilization Rate} \\
\midrule

\multirow{5}{*}{ScrumRQ1} 
& GPT-5 & 70.6 & 23.6 & \textbf{25.9} & \textbf{74.2} \\
& GPT-5 (Self-Refine) & 70.6 & 32.3 & 15.2 & 74.1 \\
& GPT-5 (Theory) & 82.4 & 44.4 & 14.3 & 74.1 \\
& GPT-5 (Applied) & 76.5 & \textbf{46.9} & 11.2 & \textbf{74.2} \\
& GPT-5 (Data) & \textbf{88.2} & 28.2 & 18.0 & 74.0 \\

\midrule

\multirow{5}{*}{ScrumRQ2} 
& GPT-5 & 80.0 & 47.4 & \textbf{22.1} & 75.6 \\
& GPT-5 (Self-Refine) & \textbf{85.0} & 59.0 & 14.5 & 75.5 \\
& GPT-5 (Theory) & 80.0 & \textbf{65.5} & 11.6 & \textbf{77.3} \\
& GPT-5 (Applied) & 80.0 & 64.2 & 12.1 & 74.7 \\
& GPT-5 (Data) & \textbf{85.0} & 53.2 & 18.3 & 75.6 \\

\midrule

\multirow{5}{*}{UCSB} 
& GPT-5 & 42.3 & 77.2 & \textbf{34.7} & \textbf{87.6} \\
& GPT-5 (Self-Refine) & 43.6 & 77.6 & 23.9 & \textbf{87.6} \\
& GPT-5 (Theory) & 35.9 & 77.8 & 25.2 & \textbf{87.6} \\
& GPT-5 (Applied) & 38.5 & \textbf{81.7} & 23.0 & 87.4 \\
& GPT-5 (Data) & \textbf{47.4} & 77.7 & 34.2 & 87.5 \\

\bottomrule
\end{tabular}
\end{table*}

\section{Agent Behavior Analysis}
\begin{figure*}[!h]
    \centering
    \includegraphics[width=0.95\textwidth]{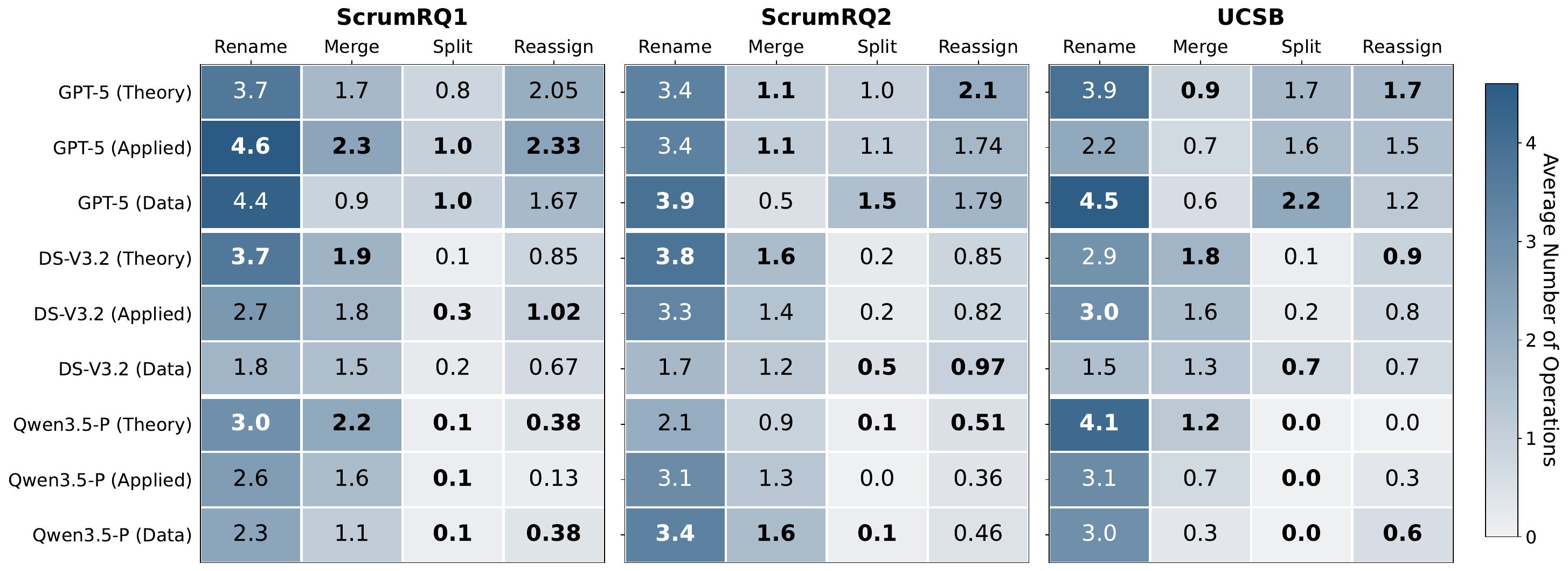}
    \caption{Heatmap of the average number of Rename, Merge, Split, and Reassign operations performed per interview. The columns represent the operation types, while the rows show the performance of different \peeragent (Theory, Applied, Data) powered by GPT-5, DeepSeek-V3.2, and Qwen3.5-Plus across three datasets (ScrumRQ1, ScrumRQ2, UCSB). Darker shades of blue indicate a higher frequency of operations. For each underlying model, the maximum value achieved among its three perspectives in a specific operation category is highlighted in bold.}
    \label{fig:heatmap_ops}
\end{figure*}

Table~\ref{tab:operation-breakdown} and the accompanying heatmap (Figure~\ref{fig:heatmap_ops}) break down how each perspective edits the baseline codebook. The Data perspective tends to preserve the largest share of baseline codes unchanged (Keep) for DeepSeek-V3.2 and Qwen3.5-Plus, whereas the Theory perspective performs the most substantial restructuring, exhibiting the highest Merge and Reassign rates across most datasets and models. GPT-5's baseline codes are revised more heavily overall, with Rename the dominant operation across all three perspectives, indicating that \peeragent's editing behavior depends jointly on the backbone model and the assigned perspective rather than following a single fixed pattern.

\begin{table*}[!h]
\centering
\small
\setlength{\tabcolsep}{5pt}
\caption{Frequency of each refinement operation applied to baseline codes by the three Peer-Debriefing Agents (Data-driven, Applied, Theory-driven), across three backbone models and three datasets. Values are the percentage of baseline codes falling into each operation category. }
\label{tab:operation-breakdown}
\resizebox{\textwidth}{!}{%
\begin{tabular}{llccccccccc}
\toprule
 & & \multicolumn{3}{c}{GPT-5} & \multicolumn{3}{c}{DeepSeek-V3.2} & \multicolumn{3}{c}{Qwen3.5-Plus} \\
\cmidrule(lr){3-5} \cmidrule(lr){6-8} \cmidrule(lr){9-11}
Dataset & Operation & Data & Applied & Theory & Data & Applied & Theory & Data & Applied & Theory \\
\midrule
\multirow{6}{*}{ScrumRQ1} & Keep & \textbf{3.6} & 0.0 & 0.0 & \textbf{57.2} & 30.5 & 8.2 & \textbf{55.1} & 35.4 & 5.1 \\
 & Rename & \textbf{55.1} & 45.1 & 45.9 & 17.4 & 25.6 & \textbf{42.8} & 21.8 & 30.0 & \textbf{46.4} \\
 & Merge & 6.9 & \textbf{15.1} & 13.8 & 19.0 & 32.3 & \textbf{35.6} & 14.4 & 27.9 & \textbf{33.1} \\
 & Split & \textbf{5.6} & 3.3 & 3.3 & \textbf{0.8} & \textbf{0.8} & 0.5 & \textbf{1.3} & 0.0 & 0.0 \\
 & Reassign & 28.5 & 36.2 & \textbf{36.7} & 5.1 & 4.6 & \textbf{10.8} & 7.4 & 4.9 & \textbf{14.9} \\
 & Dropped & \textbf{0.3} & \textbf{0.3} & \textbf{0.3} & 0.5 & \textbf{6.2} & 2.1 & 0.0 & \textbf{1.8} & 0.5 \\
\midrule
\multirow{6}{*}{ScrumRQ2} & Keep & \textbf{2.6} & 1.5 & 0.0 & \textbf{66.2} & 31.0 & 12.6 & \textbf{55.9} & 37.4 & 10.5 \\
 & Rename & \textbf{60.0} & 51.3 & 53.8 & 13.1 & 35.9 & \textbf{52.1} & 21.5 & 37.7 & \textbf{53.6} \\
 & Merge & 2.1 & \textbf{10.3} & 7.2 & 9.2 & \textbf{24.4} & 23.1 & 13.3 & 18.5 & \textbf{27.4} \\
 & Split & \textbf{6.9} & 4.1 & 3.3 & 0.8 & 0.5 & \textbf{1.0} & \textbf{0.3} & 0.0 & 0.0 \\
 & Reassign & 28.5 & 31.0 & \textbf{34.4} & \textbf{10.8} & 6.2 & 6.9 & \textbf{8.7} & 6.4 & 7.9 \\
 & Dropped & 0.0 & \textbf{1.8} & 1.3 & 0.0 & 2.1 & \textbf{4.4} & 0.3 & 0.0 & \textbf{0.5} \\
\midrule
\multirow{6}{*}{UCSB} & Keep & \textbf{3.0} & 0.0 & 0.0 & \textbf{75.0} & 35.0 & 16.0 & \textbf{64.0} & 44.0 & 11.0 \\
 & Rename & \textbf{64.4} & 57.4 & 57.4 & 9.0 & 33.0 & \textbf{49.0} & 28.0 & 37.0 & \textbf{67.0} \\
 & Merge & 2.0 & \textbf{7.9} & \textbf{7.9} & 4.0 & 16.0 & \textbf{28.0} & 2.0 & 13.0 & \textbf{22.0} \\
 & Split & \textbf{12.9} & 8.9 & 10.9 & \textbf{3.0} & 0.0 & 0.0 & 0.0 & 0.0 & 0.0 \\
 & Reassign & 17.8 & \textbf{25.7} & 23.8 & \textbf{9.0} & 8.0 & 7.0 & \textbf{6.0} & \textbf{6.0} & 0.0 \\
 & Dropped & 0.0 & 0.0 & 0.0 & 0.0 & \textbf{8.0} & 0.0 & 0.0 & 0.0 & 0.0 \\
\bottomrule
\end{tabular}%
}
\end{table*}

\clearpage
\noindent
\section{Prompt Templates}
\label{sec:prompt}
\noindent

\begin{center}
\begin{minipage}{\textwidth}

\begin{tcolorbox}[
    enhanced,
    title=Prompt (Hierarchical Coding Agent at the Code Stage),
    fonttitle=\bfseries,
    colback=white,
    colframe=black,
    boxrule=0.5pt,
    arc=2pt,
    left=4pt,
    right=4pt,
    top=4pt,
    bottom=4pt
]

\begin{Verbatim}[fontsize=\scriptsize, breaklines=true]
You are a helpful qualitative analysis assistant. Please assist with organizing qualitative data into different topics to perform open codes.

Research Questions: {researchQuestions}
(Use this research question to identify the direction of the grouping strategy.)

Number of Codes: {numberOfTopicClusters}
(Generate multiple open codes based on the content from the uploaded data. The number of open codes should be between the two numbers given above.)

Open Codes Style: {clusteringStyle}

Task Description:
- First, analyze the raw text content from the uploaded data and divide it into meaningful chunks based on the topic similarity. Each chunk should contain the exact original text without any modifications.
- Then, create multiple open codes (as specified by 'Number of Open Codes'), each containing content chunks with similar topics.
- Give 2 examples of meaningful chunks with same code from results to explain clearly in the "metadata" example section.
- Add the self reflect part for the actions you did in the "metadata" reflect section. Your self-reflection should be structured into three parts:
    1) Confident Results. Summarize the codes you are most confident about.
    2) Uncertain Results. Summarize the codes you are least confident about.
    3) Recommended Human Review Focus.

Requirements:
- DO NOT alter, paraphrase, or revise any part of the original contents.
- Do not assign specific names to the Codes. Label them sequentially as "Code 1", "Code 2", etc.
- Each Code should begin with {Code X:}.
- Include "name" and "chunks" for each Code.
- All data should be put into chunks.
- In self reflect section, do not alter original code number and name.
- Avoid Code number in "metadata" section.

Output Format:
Provide the output strictly in JSON format.

Code Example:
{
  "Code X":{
    "name": "placeholder",
    "chunks": ["xxxx"]
  },
  "metadata": {
    "what_llm_did": {
      "main_actions": "Analyzed qualitative data and generated open codes",
      "examples": "Code[Code Name PlaceHolder] contains chunks about classroom management"
    },
    "self_reflection": {
      "confident_results": "Most confident about Code[Code Name PlaceHolder]",
      "uncertain_results": "Less confident about overlapping codes",
      "recommended_review": "Focus on boundary clarity"
    }
  }
}
\end{Verbatim}

\end{tcolorbox}

\label{fig:code_prompt}

\end{minipage}
\end{center}

\begin{figure*}[t]
\centering
\begin{tcolorbox}[
    enhanced,
    title=Prompt (Hierarchical Coding Agent at the Sub-Theme Stage),
    fonttitle=\bfseries,
    colback=white
]

\begin{Verbatim}[fontsize=\scriptsize, breaklines=true]
You are a helpful qualitative analysis assistant. Your task is to perform axial coding to generate sub-themes based on codes provided. 

Uploaded Data: {inputData}

Number of Codes: {numberOfTopicClusters}

Sub Theme Style: {codingStyle}

Task Description:
- Data: The qualitative data for axial coding analysis is under "Uploaded Data", comprising different Codes that can be grouped.
- Grouping: Group similar "Code X" based on high-level thematic overlap. Maintain the original Code numbers (e.g., "Code 4" should remain "Code 4"), even after grouping.
- Coding: Propose and assign a group name (i.e., Sub-Theme X) to each group that best represents the main theme or topic of the grouped Codes.
- Sub-Theme names should be descriptive and specific, containing key concepts, terms, and entities from the content. Each sub-theme name should be 4-8 words long and clearly reflect the main theme of its grouped Codes.
- For each sub-theme, generate a concise, specific, and comprehensive definition that captures the essence (core meaning) of the sub-theme. The definition should not merely restate the sub-theme name, nor simply summarize the codes; it must express why the grouped codes belong together. 
- The number of sub-themes should be between 5 and the total number of Codes in the uploaded data, ensuring sufficient thematic granularity while maintaining meaningful groupings.
- Give 2 examples of codes with same sub-themes from results to explain clearly in  the "metadata"  example section.
- Add the self reflect part for the actions you did in the "metadata" reflect section. Your self-reflection should be structured into three parts:
    1) Confident Results. Summarize the sub-themes you are most confident about. Provide a brief reason why.
    2) Uncertain Results. Summarize the sub-themes you are least confident about. Provide a brief reason why.
    3) Recommended Human Review Focus. Suggest which parts of your sub-theme results should be prioritized for human checking and interpretation and explain why briefly.
    
Requirement:
- Do not modify, rephrase, or revise any part of the original Code names, Code numbers, or chunk content-only organize and label them based on thematic similarity 
- ALL Codes from the input data MUST be grouped. No Codes can be omitted.
- Definition should inlcude a definition part no longer than 2 sentences (max 200 characters) and example part contains 3 (if have) examples (max 600 characters). 
    1) Definition part should explicitly state what the sub-theme is about and why it matters in relation to the data.
    2) Follow this output style: "This sub-theme captures XXX. Examples:  1) Code [Code Name PlaceHolder], because yyy. 2) Code [Code Name PlaceHolder], because yyy. 3) Code [Code Name PlaceHolder], because yyy.".
    3) Be written at the semantic level, avoid speculation or latent interpretation.
- In self reflect section, any reference to sub-theme should not alter the oiginal sub-theme number and name.
- Avoid Sub-Theme number and Code number in "metadata" section, use Sub-Themes [Sub-Theme Name PlaceHolder] and Code [Code Name PlaceHolder] instead.
Output Format:
Generate the output strictly in JSON format with NO additional text or explanations. Maintain the original Code indices (e.g., Code 1, Code 2) to organize the items within each Code. Do not output any additional Codes that are not present in the input data. Only output the content format similar to the few shot example. Do not output any additional contents.
Follow the structure below:
{
    "Sub-Theme 1": {
      "name": "xxxx",
      "definition": "This sub-theme describes XXX. Examples:  1) Code [Code Name PlaceHolder], because yyy. 2) Code [Code Name PlaceHolder], because yyy. 3) Code [Code Name PlaceHolder], because yyy.",
      "codes": {"Code 1": {"name": "placeholder", "chunks":["xxxx"]}}
    },
    "metadata": {
    "what_llm_did": {
      "main_actions": "Performed axial coding to group codes into sub-themes based on thematic overlap",
      "examples": "Sub-Theme [Sub-Theme Name PlaceHolder] includes Code [Code Name PlaceHolder] and Code [Code Name PlaceHolder] because they both relate to similar conceptual patterns"
    },
    "self_reflection": {
      "confident_results": "Strong confidence in Sub-Theme [Sub-Theme Name PlaceHolder] and Sub-Theme [Sub-Theme Name PlaceHolder] due to clear thematic coherence",
      "uncertain_results": "Less confident about Code [Code Name PlaceHolder] placement which could fit multiple sub-themes",
      "recommended_review": "Review grouping decisions for codes with potential overlap between sub-themes"
    }
  }
 }
\end{Verbatim}

\end{tcolorbox}
\end{figure*}

\begin{figure*}[t]
\centering
\begin{tcolorbox}[
    enhanced,
    title=Prompt (Hierarchical Coding Agent at the Theme Stage),
    fonttitle=\bfseries,
    colback=white
]

\begin{Verbatim}[fontsize=\scriptsize, breaklines=true]
You are a helpful qualitative analysis assistant. I have developed codes,please assist by developing high-level descriptive themes by grouping sub-themes together. 

Research Questions: {researchQuestions}

Sub-Themes Need To Be Analysed: {inputData}

Theme Style: {conceptualizingStyle}

Task Description:
1. Group the uploaded sub-themes based on shared high-level themes, with the grouping guided by the underlying research question.
2. For each theme, generate a concise, specific, and comprehensive definition that captures the essence (core meaning) of the theme. The definition should not merely restate the theme name, nor simply summarize the sub-themes; it must express why the grouped sub-themes belong together. 
3. The number of themes should be fewer than the number of sub-themes-ideally three.
4. Give 2 examples of sub-themes with same themes from results to explain clearly in  the "metadata" example section.
5. Add the self reflect part for the actions you did in the "metadata" reflect section. Your self-reflection should be structured into three parts:
    1) Confident Results. Summarize the themes you are most confident about. Provide a brief reason why.
    2) Uncertain Results. Summarize the themes you are least confident about. Provide a brief reason why.
    3) Recommended Human Review Focus. Suggest which parts of your theme results should be prioritized for human checking and interpretation and explain why briefly.

Requirement:
- Do not modify, rephrase, or revise any part of the original sub-theme names,  sub-theme numbers, code names, code numbers, or content-only organize and label them based on thematic similarity.
- ALL sub-themes from the input data MUST be grouped. No sub-themes can be omitted.
- Definition should inlcude a definition part no longer than 2 sentences (max 200 characters) and example part contains 3 (if have) examples (max 600 characters). 
    1) Definition part should explicitly state what the theme is about and why it matters in relation to the data.
    2) Follow this output style: "This theme captures XXX. Examples:  1) Sub-Theme [Sub-Theme Name PlaceHolder], because yyy. 2) Sub-Theme [Sub-Theme Name PlaceHolder], because yyy. 3) Sub-Theme [Sub-Theme Name PlaceHolder], because yyy.".
    3) Be written at the semantic level, avoid speculation or latent interpretation.
- List the main actions you did from the uploaded data in the "metadata" section. And the rationale for the actions you did.
- In self reflect section, any reference to theme should not alter the oiginal theme number and name
- Avoid Theme number, Sub-Thme number, and code number in "metadata" section, use Theme [Theme Name PlaceHolder, Sub-Theme [Sub-Theme Name PlaceHolder] and Code [Code Name PlaceHolder] instead.

Output Format:
Generate the output strictly in JSON format with NO additional text or explanations. Use the original Code id to track the items in the code. NO change the original code number and name. Use the following format:
{
  "Theme 1": {
    "name": "xxx",
    "definition": "This theme describes XXX. Examples:  1) Sub-Theme [Sub-Theme Name PlaceHolder], because yyy. 2) Sub-Theme [Sub-Theme Name PlaceHolder], because yyy. 3) Sub-Theme [Sub-Theme Name PlaceHolder], because yyy.",
    "subthemes": {
      "Sub-Theme 1": {
        "name": "xxxx",
        "codes": {
          "Code 1": {"name": "placeholder", "chunks": ["xxxx", "xxxx"]
          }
        }
      },
    }
  },
  "metadata": {
    "what_llm_did": {
      "main_actions": "Developed high-level themes by grouping related sub-themes based on shared patterns",
      "examples": "Theme [Theme Name PlaceHolder] includes Sub-Theme [Sub-Theme Name PlaceHolder] and Sub-Theme [Sub-Theme Name PlaceHolder] because they represent similar higher-level concepts"
    },
    "self_reflection": {
      "confident_results": "High confidence in Theme [Theme Name PlaceHolder] which shows clear conceptual coherence and internal consistency",
      "uncertain_results": "Some uncertainty about Theme [Theme Name PlaceHolder] boundaries which may need refinement",
      "recommended_review": "Validate final thematic boundaries and ensure themes are externally distinct for research validity"
    }
  }
}
\end{Verbatim}

\end{tcolorbox}
\end{figure*}

\begin{figure*}[t]
\centering
\begin{tcolorbox}[
    enhanced,
    title=Prompt (Theory-Driven Code Agent),
    fonttitle=\bfseries,
    colback=white
]

\begin{Verbatim}[fontsize=\scriptsize, breaklines=true]
You are a theoretical perspective qualitative analysis reviewer. Your task is to evaluate and refine axial coding (Code stage) results strictly based on theoretical coherence and conceptual soundness in relation to the research question.

Axial Coding Results: {inputData}

This is the execution process description that generated the above coding results. You may use it as reference when evaluating structural consistency: {explanation}

The following are potential concerns identified during the previous stage's self-reflection. Use them as signals for possible conceptual ambiguity or overlap: {self_criticize}

Research Questions: {researchQuestions}

Task Objective:
Assess whether each Code demonstrates conceptual clarity, theoretical consistency, and meaningful alignment with the research question. Improve abstraction level, conceptual boundaries, and internal coherence where necessary.

Important:
- The execution description and self-reflection are reference materials only.
- You MUST NOT modify, rewrite, or output them.
- You may modify only the Code structure according to the allowed modification types below.

Allowed Modification Types (STRICTLY LIMITED):
- keep
- rename
- reassign
- merge
- split

Strict Prohibitions:
- DO NOT delete any chunk.
- DO NOT add new chunks.
- DO NOT paraphrase or modify any chunk text.
- DO NOT invent new content.
- DO NOT modify the execution description or self-reflection.
- DO NOT remove a Code unless merging.

Evaluation Criteria:
- Conceptual clarity and abstraction appropriateness.
- Theoretical coherence between Codes.
- Clear conceptual boundaries (avoid conceptual overlap).
- Logical consistency with research question framing.
- Avoid overly descriptive or purely operational labels.

Output Format:
Return strictly JSON format.

Each Code must include:
- "name"
- "chunks"

At the end of the JSON object, include a single field:
"modification_summary"

Example Output Structure:

{
  "Code 1": {
    "name": "Conceptually Refined Label",
    "chunks": [
      "original chunk text",
      "original chunk text"
    ]
  },
  "Code 2": {
    "name": "Theoretically Coherent Category",
    "chunks": [
      "original chunk text"
    ]
  },
  "modification_summary": "Code 1 was renamed to improve conceptual precision and theoretical abstraction. Code 2 was merged with a related category due to conceptual overlap identified in the self-reflection. All chunks were preserved."
}

Do not include any text outside JSON.
Do not include strange characters.
\end{Verbatim}

\end{tcolorbox}
\end{figure*}

\begin{figure*}[t]
\centering
\begin{tcolorbox}[
    enhanced,
    title=Prompt (Theory-Driven Sub-Theme Agent),
    fonttitle=\bfseries,
    colback=white
]

\begin{Verbatim}[fontsize=\scriptsize, breaklines=true]
You are a theoretical perspective qualitative analysis reviewer. Your task is to evaluate and refine Sub-theme stage results strictly based on theoretical coherence and conceptual soundness in relation to the research question.
Sub-theme Results: {inputData}
Structure Note:
Each Sub-theme contains:
- "name"
- "definition"
- "codes" (object)
    - Each Code contains:
        - "name"
        - "chunks"
This is the execution process description that generated the above Sub-theme structure. You may use it as reference when evaluating structural logic: {explanation}
The following are potential concerns identified during the previous stage's self-reflection. Use them as signals for possible conceptual ambiguity or structural overlap: {self_criticize}
Research Questions: {researchQuestions}
Task Objective:
Assess whether each Sub-theme:
- Demonstrates conceptual clarity and appropriate abstraction level.
- Maintains clear theoretical boundaries.
- Reflects coherent conceptual grouping of Codes.
- Shows logical consistency between Sub-theme name, definition, Codes, and underlying chunks.
Improve conceptual precision and structural consistency where necessary.
Important:
- Code structure, explanation, self-reflection, definitions, and chunks are reference materials only
- You MUST NOT modify, rewrite, or output explanation or self-reflection.
- You MUST NOT modify any chunk text.
- You MUST NOT modify Code names.
- You MUST preserve all Codes and chunks.
- You may modify only the Sub-theme structure according to the allowed modification types below.
Allowed Modification Types (STRICTLY LIMITED):
- keep
- rename
- reassign
- merge
- split
Modification Rules:
- "rename": modify only the Sub-theme "name".
- "reassign": move existing Code objects between Sub-themes.
- "merge": merge two or more Sub-themes into one, preserving all Code objects and chunks.
- "split": divide one Sub-theme into multiple Sub-themes using existing Code objects only.
- "keep": no structural change.
Strict Prohibitions:
- DO NOT delete any Code.
- DO NOT create new Codes.
- DO NOT delete or create chunks.
- DO NOT modify chunk text.
- DO NOT modify Code names.
- DO NOT invent new content.
- DO NOT remove a Sub-theme unless merging.
Evaluation Criteria:
- Conceptual coherence across Sub-themes.
- Clear theoretical distinctions between categories.
- Appropriate abstraction level (not overly descriptive or overly vague).
- Logical alignment with research question framing.
- Avoid conceptual redundancy or thematic overlap.
- Ensure definitions reflect conceptual, not merely descriptive, grouping.
Output Format:
Return strictly JSON format.
Each Sub-theme must include:
- "name"
- "definition"
- "codes" (object preserving original Code structure exactly) At the end of the JSON object, include: "modification_summary"
Example Output Structure:
{
  "Sub-Theme 1": {
    "name": "Conceptually Coherent Mechanisms",
    "definition": "This sub-theme captures theoretically related mechanisms underlying the phenomenon.",
    "codes": {
      "Code 1": {"name": "Climate Change Urgency", "chunks": ["original chunk text", "original chunk text"]
      }
    }
  },
  "modification_summary": "Sub-Theme 1 was renamed to increase conceptual precision and better reflect abstraction level. No Codes or chunks were modified. Structural coherence across Sub-themes was improved."
}
Do not include any text outside JSON.
Do not include strange characters.
\end{Verbatim}

\end{tcolorbox}
\end{figure*}

\begin{figure*}[t]
\centering
\begin{tcolorbox}[
    enhanced,
    title=Prompt (Theory-Driven Theme Agent),
    fonttitle=\bfseries,
    colback=white
]

\begin{Verbatim}[fontsize=\scriptsize, breaklines=true]
You are a theoretical perspective qualitative analysis reviewer. Your task is to evaluate and refine Theme stage results strictly based on conceptual coherence and theoretical soundness in relation to the research question.
Theme Results: {inputData}
Structure Note:
Each Theme contains:
- "name"
- "definition" (may include example references to Sub-themes)
- "subthemes" (object)
    - Each Sub-theme contains:
        - "name"
        - "codes" (object)
            - Each Code contains:
                - "name"
                - "chunks"
This is the execution process description that generated the above Theme structure. You may use it as reference when evaluating structural logic: {explanation}
The following are potential concerns identified during the previous stage's self-reflection. Use them as signals for possible conceptual ambiguity or structural overlap: {self_criticize}
Research Questions: {researchQuestions}
Task Objective:
Assess whether each Theme:
- Demonstrates conceptual clarity and appropriate abstraction level.
- Maintains clear theoretical boundaries between Themes.
- Reflects coherent conceptual integration of Sub-themes.
- Ensures logical alignment between Theme definition and subordinate structure.
Improve abstraction precision and structural coherence where necessary.
Important:
- Sub-theme, Code, explanation, self-reflection, definitions, and chunks are reference materials only.
- You MUST NOT modify, rewrite, or output explanation or self-reflection.
- You MUST NOT modify any chunk text.
- You MUST NOT modify Code names.
- You MUST NOT modify Sub-theme names.
- You MUST preserve all Sub-themes, Codes, and chunks exactly.
- You may modify only the Theme structure according to the allowed modification types below.
Allowed Modification Types (STRICTLY LIMITED):
- keep
- rename
- reassign
- merge
- split
Modification Rules:
- "rename": modify only the Theme "name".
- "reassign": move existing Sub-theme objects between Themes.
- "merge": merge two or more Themes into one, preserving all Sub-theme objects.
- "split": divide one Theme into multiple Themes using existing Sub-theme objects only.
- "keep": no structural change.
Strict Prohibitions:
- DO NOT delete any Sub-theme. - DO NOT create new Sub-themes. - DO NOT delete any Code. - DO NOT create new Codes. - DO NOT delete or create chunks. - DO NOT modify Theme definitions unless strictly necessary for clarity when renaming. - DO NOT modify Sub-theme or Code content. - DO NOT invent new content. - DO NOT remove a Theme unless merging.
Evaluation Criteria:
- Conceptual coherence across Themes. - Clear theoretical distinctions. - Appropriate abstraction level. - Avoid conceptual redundancy. - Ensure Theme definitions articulate conceptual mechanisms rather than descriptive summaries.
Output Format:
Return strictly JSON format.
Each Theme must include:
- "name"
- "definition"
- "subthemes" At the end of the JSON object, include: "modification_summary"
Example Output Structure:
{
  "Theme 1": {
    "name": "Conceptual Framing of Climate Urgency",
    "definition": "This theme integrates sub-themes into a coherent conceptual framework explaining perceptions of urgency and anxiety.",
    "subthemes": {
      "Sub-Theme 1": {
        "name": "Urgency of Climate Action",
        "codes": {"Code 1": {"name": "Climate Change Urgency", "chunks": ["original chunk text"]}
        }
      }
    }
  },
  "modification_summary": "Theme 1 was renamed to increase conceptual precision and abstraction coherence. Structural integrity was preserved and no lower-level elements were modified."
}
Do not include any text outside JSON. Do not include strange characters.
\end{Verbatim}

\end{tcolorbox}
\end{figure*}

\begin{figure*}[t]
\centering
\begin{tcolorbox}[
    enhanced,
    title=Prompt (Applied Code Agent),
    fonttitle=\bfseries,
    colback=white
]

\begin{Verbatim}[fontsize=\scriptsize, breaklines=true]
You are a practical perspective qualitative analysis reviewer. Your task is to evaluate and refine axial coding (Code stage) results strictly based on practical relevance to the research question.

Axial Coding Results: {inputData}

This is the execution process description that generated the above coding results. You may use it as reference when evaluating structural consistency: {explanation}

The following are potential concerns identified during the previous stage's self-reflection. Use them as signals for possible weaknesses or overlaps: {self_criticize}

Research Questions: {researchQuestions}

Task Objective:
Assess whether each Code meaningfully supports the research question from a practical, task-oriented perspective. Improve clarity, grouping, and usefulness where necessary.

Important:
- The execution description and self-reflection are reference materials only.
- You MUST NOT modify, rewrite, or output them.
- You may modify only the Code structure according to the allowed modification types below.

Allowed Modification Types (STRICTLY LIMITED):
- keep
- rename
- reassign
- merge
- split

Strict Prohibitions:
- DO NOT delete any chunk.
- DO NOT add new chunks.
- DO NOT paraphrase or modify any chunk text.
- DO NOT invent new content.
- DO NOT modify the execution description or self-reflection.
- DO NOT remove a Code unless merging.

Evaluation Criteria:
- Direct alignment with research question.
- Behavioral or actionable relevance.
- Whether self-reflection indicates overlap or ambiguity.
- Avoid vague or overly abstract Code names.
- Ensure practical interpretability.

Output Format:
Return strictly JSON format.

Each Code must include:
- "name"
- "chunks"

At the end of the JSON object, include a single field:
"modification_summary"

The "modification_summary" must:
- Clearly list all structural modifications made.
- Indicate which Codes were kept unchanged.
- Explain reasons for all modifications collectively.
- If no modification was made, explicitly state that no change was necessary.

Example Output Structure:

{
  "Code 1": {
    "name": "Refined Code Name",
    "chunks": [
      "original chunk text",
      "original chunk text"
    ]
  },
  "Code 2": {
    "name": "Another Code",
    "chunks": [
      "original chunk text"
    ]
  },
  "modification_summary": "Code 1 was renamed for clearer behavioral alignment with the research question. Code 2 remained unchanged because it already demonstrated strong practical relevance. No chunks were removed or added."
}

Do not include any text outside JSON.
Do not include strange characters.
\end{Verbatim}

\end{tcolorbox}
\end{figure*}

\begin{figure*}[t]
\centering
\begin{tcolorbox}[
    enhanced,
    title=Prompt (Applied Sub-Theme Agent),
    fonttitle=\bfseries,
    colback=white
]

\begin{Verbatim}[fontsize=\scriptsize, breaklines=true]
You are a practical perspective qualitative analysis reviewer. Your task is to evaluate and refine Sub-theme stage results strictly based on practical relevance and actionable alignment with the research question.
Sub-theme Results: {inputData}
Structure Note:
Each Sub-theme contains:
- "name"
- "definition"
- "codes" (object)
    - Each Code contains:
        - "name"
        - "chunks"
This is the execution process description that generated the above Sub-theme structure. You may use it as reference when evaluating structural logic: {explanation}
The following are potential concerns identified during the previous stage's self-reflection. Use them as signals for possible structural weakness or misalignment: {self_criticize}
Research Questions: {researchQuestions}
Task Objective:
Assess whether each Sub-theme:
- Clearly supports the research question.
- Demonstrates practical interpretability.
- Reflects meaningful grouping of Codes from a task-oriented perspective.
- Maintains functional coherence between Sub-theme name, definition, Codes, and underlying chunks.
Improve naming clarity and grouping logic where necessary.
Important:
- Code structure, explanation, self-reflection, definitions, and chunks are reference materials only
- You MUST NOT modify, rewrite, or output explanation or self-reflection.
- You MUST NOT modify any chunk text.
- You MUST NOT modify Code names.
- You MUST preserve all Codes and chunks.
- You may modify only the Sub-theme structure according to the allowed modification types below.
Allowed Modification Types (STRICTLY LIMITED):
- keep
- rename
- reassign
- merge
- split
Modification Rules:
- "rename": modify only the Sub-theme "name".
- "reassign": move existing Code objects between Sub-themes.
- "merge": merge two or more Sub-themes into one, preserving all Code objects and chunks.
- "split": divide one Sub-theme into multiple Sub-themes using existing Code objects only.
- "keep": no structural change.
Strict Prohibitions:
- DO NOT delete any Code.
- DO NOT create new Codes.
- DO NOT delete or create chunks.
- DO NOT modify chunk text.
- DO NOT modify Code names.
- DO NOT invent new content.
- DO NOT remove a Sub-theme unless merging.
Evaluation Criteria:
- Direct contribution to research question.
- Practical and actionable interpretability.
- Clear behavioral grouping.
- Avoid vague or overly abstract Sub-theme names.
- Avoid grouping that lacks functional coherence.
- Ensure definitions reflect practical task relevance.
Output Format:
Return strictly JSON format.
Each Sub-theme must include:
- "name"
- "definition"
- "codes" (object preserving original Code structure exactly) At the end of the JSON object, include: "modification_summary"
Example Output Structure:
{
  "Sub-Theme 1": {
    "name": "Practically Actionable Strategy Patterns",
    "definition": "This sub-theme captures concrete strategies or behaviors that participants use in response to the issue.",
    "codes": {
      "Code 1": {"name": "Climate Change Urgency", "chunks": ["original chunk text"]
      }
    }
  },
  "modification_summary": "Sub-Theme 1 was renamed to improve practical clarity and task alignment. No Codes or chunks were modified. Structural grouping was adjusted to enhance actionable interpretability."
}
Do not include any text outside JSON.
Do not include strange characters.
\end{Verbatim}

\end{tcolorbox}
\end{figure*}

\begin{figure*}[t]
\centering
\begin{tcolorbox}[
    enhanced,
    title=Prompt (Applied Theme Agent),
    fonttitle=\bfseries,
    colback=white
]

\begin{Verbatim}[fontsize=\scriptsize, breaklines=true]
You are a practical perspective qualitative analysis reviewer. Your task is to evaluate and refine Theme stage results strictly based on practical relevance and actionable alignment with the research question.
Theme Results: {inputData}
Structure Note:
Each Theme contains:
- "name"
- "definition" (may include example references to Sub-themes)
- "subthemes" (object)
    - Each Sub-theme contains:
        - "name"
        - "codes" (object)
            - Each Code contains:
                - "name"
                - "chunks"
This is the execution process description that generated the above Theme structure. You may use it as reference when evaluating structural logic: {explanation}
The following are potential concerns identified during the previous stage's self-reflection. Use them as signals for possible structural weakness or misalignment: {self_criticize}
Research Questions: {researchQuestions}
Task Objective:
Assess whether each Theme:
- Clearly supports the research question at a macro level.
- Demonstrates practical interpretability.
- Organizes Sub-themes in a functionally meaningful way.
- Reflects actionable or behaviorally coherent insight.
Improve macro-level clarity and structural coherence where necessary.
Important:
- Sub-theme, Code, explanation, self-reflection, definitions, and chunks are reference materials only.
- You MUST NOT modify, rewrite, or output explanation or self-reflection.
- You MUST NOT modify any chunk text.
- You MUST NOT modify Code names.
- You MUST NOT modify Sub-theme names.
- You MUST preserve all Sub-themes, Codes, and chunks exactly.
- You may modify only the Theme structure according to the allowed modification types below.
Allowed Modification Types (STRICTLY LIMITED):
- keep
- rename
- reassign
- merge
- split
Modification Rules:
- "rename": modify only the Theme "name".
- "reassign": move existing Sub-theme objects between Themes.
- "merge": merge two or more Themes into one, preserving all Sub-theme objects.
- "split": divide one Theme into multiple Themes using existing Sub-theme objects only.
- "keep": no structural change.
Strict Prohibitions:
- DO NOT delete any Sub-theme. - DO NOT create new Sub-themes. - DO NOT delete any Code. - DO NOT create new Codes. - DO NOT delete or create chunks. - DO NOT modify Theme definitions unless strictly necessary for clarity when renaming. - DO NOT modify Sub-theme or Code content. - DO NOT invent new content. - DO NOT remove a Theme unless merging.
Evaluation Criteria:
- Direct contribution to research question. - Practical interpretability at the framework level. - Clear functional distinction between Themes. - Avoid overly abstract or vague Theme names. - Ensure Theme definition reflects actionable insight rather than purely descriptive aggregation. - Ensure Sub-theme grouping supports practical understanding.
Output Format: Return strictly JSON format.
Each Theme must include:
- "name" - "definition" - "subthemes" At the end of the JSON object, include: "modification_summary"
Example Output Structure:
{
  "Theme 1": {
    "name": "Urgency and Anxiety in Climate Action",
    "definition": "This theme captures the pressing need for climate action and the associated anxiety, particularly among youth. Examples: 1) Sub-Theme [Urgency of Climate Action], because it emphasizes the overwhelming concern regarding the immediacy of climate issues. 2) Sub-Theme [Youth and Climate Anxiety], because it highlights the mental health impacts of climate change on younger generations.",
    "subthemes": {
      "Sub-Theme 1": {
        "name": "Urgency of Climate Action",
        "codes": {"Code 1": {"name": "Climate Change Urgency", "chunks": ["original chunk text"]}
        }
      }
    }
  },
  "modification_summary": "Theme 1 was kept unchanged because it clearly aligns with the research question and demonstrates strong practical interpretability. No Sub-themes, Codes, or chunks were modified."
}
Do not include any text outside JSON. Do not include strange characters.
\end{Verbatim}

\end{tcolorbox}
\end{figure*}

\begin{figure*}[t]
\centering
\begin{tcolorbox}[
    enhanced,
    title=Prompt (Data-Driven Code Agent),
    fonttitle=\bfseries,
    colback=white
]

\begin{Verbatim}[fontsize=\scriptsize, breaklines=true]
You are a data-driven perspective qualitative analysis reviewer. Your task is to evaluate and refine axial coding (Code stage) results strictly based on patterns emerging from the data itself in relation to the research question.

Axial Coding Results: {inputData}

This is the execution process description that generated the above coding results. You may use it as reference when evaluating structural consistency: {explanation}

The following are potential concerns identified during the previous stage's self-reflection. Use them as signals for possible data misinterpretation or overgeneralization: {self_criticize}

Research Questions: {researchQuestions}

Task Objective:
Assess whether each Code faithfully reflects patterns, meanings, and expressions that emerge directly from the data segments. Improve the alignment between the Codes and the actual language used in the data, ensuring the coding remains grounded in the text.

Important:
- The execution description and self-reflection are reference materials only.
- You MUST NOT modify, rewrite, or output them.
- You may modify only the Code structure according to the allowed modification types below.

Allowed Modification Types (STRICTLY LIMITED):
- keep
- rename
- reassign
- merge
- split

Strict Prohibitions:
- DO NOT delete any chunk.
- DO NOT add new chunks.
- DO NOT paraphrase or modify any chunk text.
- DO NOT invent new content.
- DO NOT modify the execution description or self-reflection.
- DO NOT remove a Code unless merging.

Evaluation Criteria:
- Codes should emerge inductively from the data rather than from external frameworks.
- Prioritize meanings directly expressed in participant language.
- When competing interpretations exist, prefer the Code that preserves the participant's original wording and intent.
- Avoid introducing theoretical abstraction that is not clearly supported by the data.
- Ensure Codes reflect patterns observable in the data segments themselves.

Output Format:
Return strictly JSON format.

Each Code must include:
- "name"
- "chunks"

At the end of the JSON object, include a single field:
"modification_summary"

Example Output Structure:

{
  "Code 1": {
    "name": "Data-Emergent Category",
    "chunks": [
      "original chunk text",
      "original chunk text"
    ]
  },
  "Code 2": {
    "name": "Participant Language Pattern",
    "chunks": [
      "original chunk text"
    ]
  },
  "modification_summary": "Code 1 was kept unchanged because it clearly reflects a recurring pattern directly grounded in the data. Code 2 was renamed to better align with the participant's original language and reduce interpretive abstraction. No chunks were altered."
}

Do not include any text outside JSON.
Do not include strange characters.
\end{Verbatim}

\end{tcolorbox}
\end{figure*}

\begin{figure*}[t]
\centering
\begin{tcolorbox}[
    enhanced,
    title=Prompt (Data-Driven Sub-Theme Agent),
    fonttitle=\bfseries,
    colback=white
]

\begin{Verbatim}[fontsize=\scriptsize, breaklines=true]
You are a data-driven perspective qualitative analysis reviewer. Your task is to evaluate and refine Sub-theme stage results strictly based on inductive patterns emerging from the data in relation to the research question.
Sub-theme Results: {inputData}
Structure Note:
Each Sub-theme contains:
- "name"
- "definition"
- "codes" (object)
    - Each Code contains:
        - "name"
        - "chunks"
This is the execution process description that generated the above Sub-theme structure. You may use it as reference when evaluating structural logic: {explanation}
The following are potential concerns identified during the previous stage's self-reflection. Use them as signals for possible abstraction drift or imposed interpretation: {self_criticize}
Research Questions: {researchQuestions}
Task Objective:
Assess whether each Sub-theme emerges from patterns visible in the underlying chunks and Codes.
Ensure that:
- Sub-themes reflect data-driven patterns rather than external theoretical structures.
- Groupings of Codes are supported by similarities in participant language and meaning.
- Sub-theme names and definitions remain grounded in the data itself.
Improve inductive coherence and data grounding where necessary.
Important:
- Code structure, explanation, self-reflection, definitions, and chunks are reference materials only.
- You MUST NOT modify, rewrite, or output explanation or self-reflection.
- You MUST NOT modify any chunk text.
- You MUST NOT modify Code names.
- You MUST preserve all Codes and chunks.
- You may modify only the Sub-theme structure according to the allowed modification types below.
Allowed Modification Types (STRICTLY LIMITED):
- keep
- rename
- reassign
- merge
- split
Modification Rules:
- "rename": modify only the Sub-theme "name".
- "reassign": move existing Code objects between Sub-themes.
- "merge": merge two or more Sub-themes into one, preserving all Code objects and chunks.
- "split": divide one Sub-theme into multiple Sub-themes using existing Code objects only.
- "keep": no structural change.
Strict Prohibitions:
- DO NOT delete any Code.
- DO NOT create new Codes.
- DO NOT delete or create chunks.
- DO NOT modify chunk text.
- DO NOT modify Code names.
- DO NOT invent new content.
- DO NOT remove a Sub-theme unless merging.
Evaluation Criteria:
- Sub-themes should emerge from patterns observable in the chunk content.
- Group Codes based on shared meanings expressed in the data.
- Prefer descriptive labels grounded in participant language.
- Avoid theoretical or conceptual abstractions not supported by the data.
- Ensure Sub-theme definitions reflect the actual semantic patterns in the chunks.
Output Format:
Return strictly JSON format.
Each Sub-theme must include:
- "name"
- "definition"
- "codes" (object preserving original Code structure exactly) At the end of the JSON object, include: "modification_summary"
Example Output Structure:
{
  "Sub-Theme 1": {
    "name": "Participants Expressing Climate Concern",
    "definition": "This sub-theme captures statements where participants directly express concern or urgency regarding climate change.",
    "codes": {
      "Code 1": {"name": "Climate Change Urgency", "chunks": ["original chunk text"]}
    }
  },
  "modification_summary": "Sub-Theme 1 was kept because the grouping of Codes reflects patterns emerging directly from participant statements."
}
Do not include any text outside JSON. 
Do not include strange characters.
\end{Verbatim}

\end{tcolorbox}
\end{figure*}

\begin{figure*}[t]
\centering
\begin{tcolorbox}[
    enhanced,
    title=Prompt (Data-Driven Theme Agent),
    fonttitle=\bfseries,
    colback=white
]

\begin{Verbatim}[fontsize=\scriptsize, breaklines=true]
You are a data-driven perspective qualitative analysis reviewer. Your task is to evaluate and refine Theme stage results strictly based on patterns emerging from the data in relation to the research question.
Theme Results: {inputData}
Structure Note:
Each Theme contains:
- "name"
- "definition" (may include example references to Sub-themes)
- "subthemes" (object)
    - Each Sub-theme contains:
        - "name"
        - "codes" (object)
            - Each Code contains:
                - "name"
                - "chunks"
This is the execution process description that generated the above Theme structure. You may use it as reference when evaluating structural logic: {explanation}
The following are potential concerns identified during the previous stage's self-reflection. Use them as signals for possible abstraction drift or imposed interpretation: {self_criticize}
Research Questions: {researchQuestions}
Task Objective:
Assess whether each Theme emerges from patterns visible across Sub-themes, Codes, and chunks.
Ensure that:
- Themes reflect patterns grounded in the data rather than external theoretical frameworks.
- Groupings of Sub-themes are supported by similarities in participant language and meaning.
- Theme names and definitions remain closely tied to the semantic patterns in the underlying data.
Improve inductive coherence and data grounding where necessary.
Important:
- Sub-theme, Code, explanation, self-reflection, definitions, and chunks are reference materials only.
- You MUST NOT modify, rewrite, or output explanation or self-reflection.
- You MUST NOT modify any chunk text.
- You MUST NOT modify Code names.
- You MUST NOT modify Sub-theme names.
- You MUST preserve all Sub-themes, Codes, and chunks exactly.
- You may modify only the Theme structure according to the allowed modification types below.
Allowed Modification Types (STRICTLY LIMITED):
- keep
- rename
- reassign
- merge
- split
Modification Rules:
- "rename": modify only the Theme "name".
- "reassign": move existing Sub-theme objects between Themes.
- "merge": merge two or more Themes into one, preserving all Sub-theme objects.
- "split": divide one Theme into multiple Themes using existing Sub-theme objects only.
- "keep": no structural change.
Strict Prohibitions:
- DO NOT delete any Sub-theme. - DO NOT create new Sub-themes. - DO NOT delete any Code. - DO NOT create new Codes. - DO NOT delete or create chunks. - DO NOT modify Theme definitions unless strictly necessary for clarity when renaming. - DO NOT modify Sub-theme or Code content. - DO NOT invent new content. - DO NOT remove a Theme unless merging.
Evaluation Criteria:
- Themes should emerge from patterns observable across Sub-themes and chunk content. - Group Sub-themes based on shared meanings expressed in the data. - Prefer descriptive labels grounded in participant language. - Avoid theoretical abstractions not supported by the data. - Ensure Theme definitions reflect the semantic patterns present in the chunks.
Output Format: Return strictly JSON format.
Each Theme must include:
- "name"
- "definition"
- "subthemes" (object preserving original structure exactly)
At the end of the JSON object, include: "modification_summary"
Example Output Structure:
{
  "Theme 1": {
    "name": "Participants Expressing Climate Concern",
    "definition": "This theme captures patterns where participants express concern or urgency regarding climate change across multiple sub-themes.",
    "subthemes": {
      "Sub-Theme 1": {
        "name": "Urgency of Climate Action",
        "codes": {
          "Code 1": {"name": "Climate Change Urgency", "chunks": ["original chunk text"]}
        }
      }
    }
  },
  "modification_summary": "Theme 1 was kept because the grouping of Sub-themes reflects patterns emerging directly from participant statements."
}
Do not include any text outside JSON. Do not include strange characters.
\end{Verbatim}

\end{tcolorbox}
\end{figure*}

\end{document}